\documentclass[letterpaper, 10 pt, journal, twoside]{ieeetran}

\IEEEoverridecommandlockouts    

\usepackage{cite}
\usepackage{algorithmic}
\usepackage{textcomp}
\def\BibTeX{{\rm B\kern-.05em{\sc i\kern-.025em b}\kern-.08em
    T\kern-.1667em\lower.7ex\hbox{E}\kern-.125emX}}

\usepackage{graphicx}
\usepackage{caption}
\usepackage{times} 
\usepackage{amsmath} 
\usepackage{amssymb}  
\usepackage{amsfonts} 
\usepackage{mathrsfs}
\usepackage[mathscr]{euscript}
\usepackage{times}
\usepackage[T1]{fontenc}
\usepackage[colorlinks=true,urlcolor=BrickRed]{hyperref}
\usepackage{balance}

\newcommand{\transpose}{\mathsf{T}}

\newcommand{\blue}[1]{\textcolor{black}{#1}}

\usepackage[dvipsnames]{xcolor} 

\title{Tensegrity Robot Proprioceptive State Estimation with Geometric~Constraints\\
}

\author{Wenzhe Tong, Tzu-Yuan Lin, Jonathan Mi, Yicheng Jiang, Maani Ghaffari, and Xiaonan Huang
    \thanks{Manuscript received: September, 27, 2024; Revised December, 13, 2024; Accepted February, 8, 2025.}
    \thanks{This paper was recommended for publication by Editor Cecilia Laschi upon evaluation of the Associate Editor and Reviewers' comments.
    This work was supported by the startup fund from the Robotics Department at the University of Michigan.} 
    \thanks{$^{1}$The authors are with the Robotics Department, University of Michigan, Ann Arbor, MI, 48109, USA.
        {\tt\footnotesize\{wenzhet, tzuyuan, jjomi, valeska, maanigj, xiaonanh\}@umich.edu}}
    \thanks{Digital Object Identifier (DOI): see top of this page.}
}

\begin{document}

\maketitle

\begin{abstract}
Tensegrity robots, characterized by a synergistic assembly of rigid rods and elastic cables, form robust structures that are resistant to impacts. However, this design introduces complexities in kinematics and dynamics, complicating control and state estimation. This work presents a novel proprioceptive state estimator for tensegrity robots. The estimator initially uses the geometric constraints of 3-bar prism tensegrity structures, combined with IMU and motor encoder measurements, to reconstruct the robot's shape and orientation. It then employs a contact-aided invariant extended Kalman filter with forward kinematics to estimate the global position and orientation of the tensegrity robot.
The state estimator's accuracy is assessed against ground truth data in both simulated environments and real-world tensegrity robot applications. It achieves an average drift percentage of 4.2\%, comparable to the state estimation performance of traditional rigid robots. This state estimator advances the state-of-the-art in tensegrity robot state estimation and has the potential to run in real-time using onboard sensors, paving the way for full autonomy of tensegrity robots in unstructured environments.

\end{abstract}


\section{Introduction}
\IEEEPARstart{T}{ensegrity} robots are composed of rigid rods suspended by a network of cables or tendons. \blue{This tensional integrity structural design enhances the robot's resilience, allowing it to absorb impacts\cite{skelton2009tensegrity, shah2022tensegrity}, such as surviving drops from height. Furthermore, it provides shape-morphing capabilities that improve locomotion in challenging environments (e.g., navigating through confined spaces) and enables the structure to collapse into smaller volumes, making transportation and deployment more efficient.} Most existing research on tensegrity robots has predominantly focused on their design\cite{sabelhaus2014hardware, sabelhaus2015system, vespignani2018design, mi2024design}, control\cite{paul2006design, kim2014rapid, kim2015robust}, and dynamics\cite{paul2006design, shah2022tensegrity}, with relatively limited attention given to the challenge of state estimation. The inherent flexibility and deformable structure of the design, combined with their unique rolling maneuver, introduce significant challenges in accurately reconstructing the robots' shape and estimating their pose.

\begin{figure}[t]
    \centering
    \includegraphics[width=0.8\linewidth]{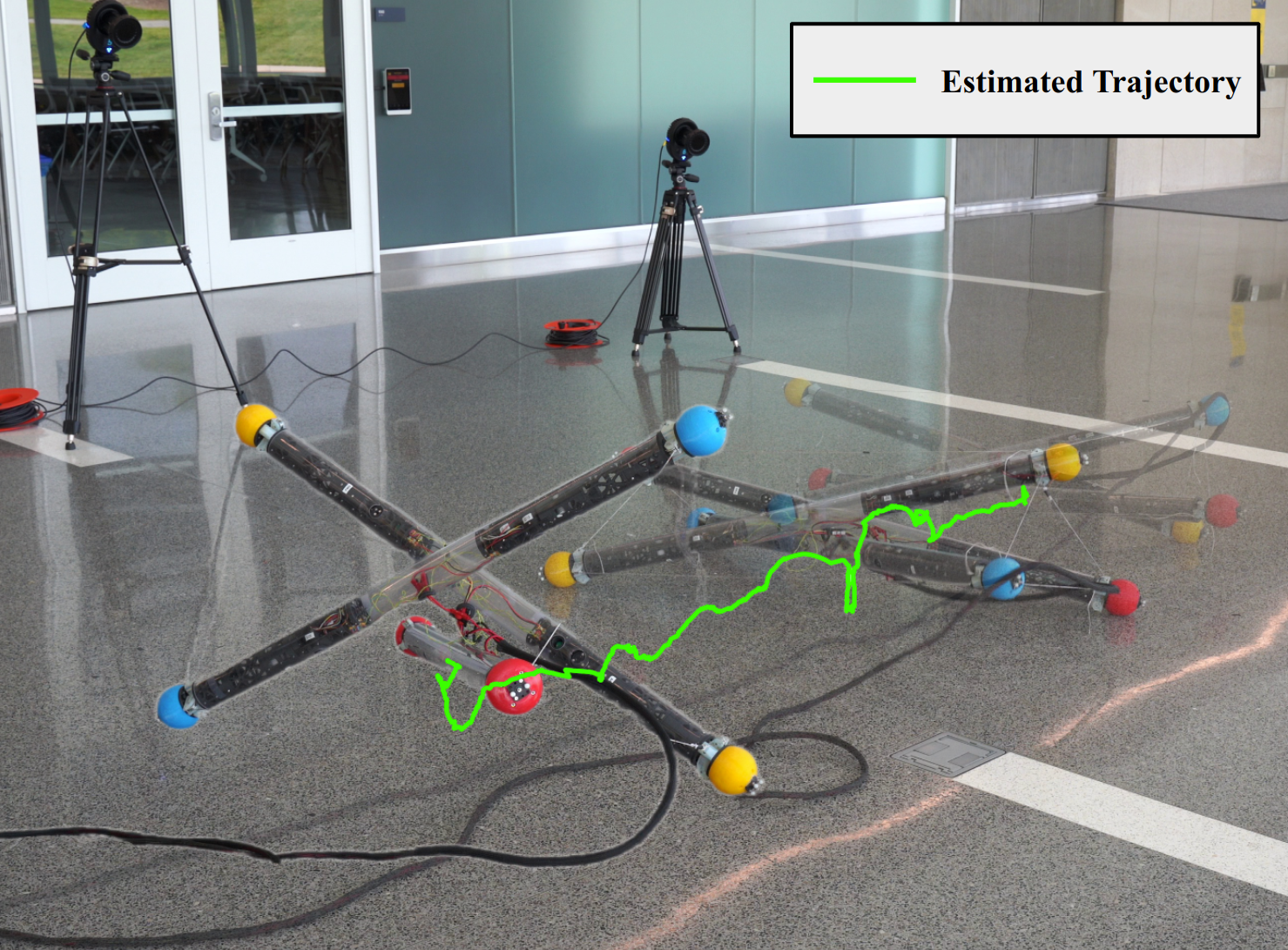}
    \caption{Tensegrity robot state estimation experiment setup, the robot is teleoperated to roll on flat ground.}
    \label{fig:tensegrity-locomotion-trajectory-legend}
    \vspace{-6mm}
\end{figure}

Autonomous robots with motion planning and feedback control capabilities rely extensively on accurate state estimation, i.e., position, orientation, and velocity. Exteroceptive sensors, e.g., cameras and LiDARs, are commonly used to provide perceptual data for global localization and navigation\cite{campos2021orb, shan2020lio}. However, these sensors typically operate at low frequencies (10-30Hz) and are susceptible to illumination change, featureless scenarios\cite{li2020dxslam, lin2023proprioceptive}, and motion blurs\cite{xu2024customizable}. \blue{Proprioceptive sensors, such as IMUs and joint encoders, offer real-time high-frequency (50-1000Hz) measurements, which can capture highly dynamic motion, such as rolling transitions for tensegrity robots. These sensors are also essential for high-frequency feedback controllers and state estimators and offer greater reliability as they operate independently of environmental conditions, ensuring resilience against external disturbances.}


Many existing tensegrity robots are often sensor-free \cite{chen2017soft, hirai2013active, paul2006design, rieffel2018adaptive, kim2014rapid}, making them incapable of performing independent state estimation, or they rely on external sensors to achieve this functionality. Only a few tensegrity robots\cite{lu20226n, johnson2022sensor, li2021shape} can estimate their shape using onboard sensors, but they lack the capability for pose estimation.
Conventional vision-based pose estimation techniques\cite{qin2018vins, campos2021orb} prove challenging due to the deformable nature of the tensegrity robots and issues related to self-occlusion. This underscores the need to develop robust, high-frequency proprioceptive state estimators that utilize onboard sensors. To the best of our knowledge, no tensegrity robot currently possesses the capability for state estimation relying solely on proprioceptive sensors, which significantly hinders the application of tensegrity robots in real-world exploration tasks.

In this paper, we present a proprioceptive state estimator for the tensegrity robot using an invariant extended Kalman filter (InEKF). Our approach first estimates the robot's shape in the body frame as kinematics information and subsequently integrates the inertial measurements to estimate the robot's position and orientation. The proposed method is evaluated quantitatively in both Mujoco simulation and real-world settings, where the tensegrity robot is locomoted using pre-computed gaits. The estimated poses are compared against ground truth data obtained from the simulation or a motion capture system, achieving an average of 4.2\% final drift percentage, which is comparable to the reported performance metrics for wheeled robots (3.8\%), legged robots (1.65 \%) and full-size vehicles (3.18\%) as noted in \cite{lin2023proprioceptive}.

The contributions of this paper are: 

\begin{enumerate}
    \item We introduce the first proprioceptive InEKF state estimator capable of estimating both the shape and pose of the tensegrity robot, featuring a measurement model tailored to the kinematics of tensegrity robots. 
    \item We incorporate the geometric properties of the 3-bar tensegrity robot as constraints in the shape reconstruction process via constrained optimization. 
    \item We validate the proposed method through extensive simulation and real-world experiments using the 3-bar tensegrity robot performing rolling maneuvers over varied terrains.
    \item The code can be accessed at: {\footnotesize \url{https://github.com/Jonathan-Twz/tensegrity-robot-state-estimator}}
\end{enumerate}

\section{Related Work}

Proprioceptive state estimation problems for traditional rigid robots are well-studied. 
For legged robots, the initial proprioceptive state estimation methods were based on leg kinematics, known as legged odometry (LO)\cite{roston1991dead, camurri2020pronto}. Legged odometry assumes the foot contact position is static in the world frame, and the robot's state is estimated using the motion of the legs, including position and orientation, through forward kinematics based on joint angles and foot contacts. More recent work has focused on incorporating IMU data with legged odometry (LIO) using Kalman Filter (KF)\cite{li2013high}. Bloesch et al. \cite{bloesch2013state} proposed estimating robot pose, velocity, and IMU biases using body IMU, joint encoders, and foot contact sensors.
Hartley et al.~\cite{hartley2020contact} designed a contact-aided Invariant Extended Kalman Filter (InEKF) for legged robot state estimation. The IMU strapdown model was used for propagation, while foot contact positions were augmented in the state during the correction step using forward kinematics. The contact-aided InEKF improved orientation estimation convergence and overall performance on the bipedal Cassie robot.

Wheeled robots and autonomous vehicles, on the other hand, utilize data from IMU and wheel encoders to deliver robust state estimates even under collisions or slippage conditions \cite{yu2023fully, xiong2019imu}. \blue{Legged and wheeled robots benefit from deterministic kinematics, where control inputs are directly mapped to body velocity. However, tensegrity robots pose a unique challenge, as their kinematics are inherently linked to continuously evolving structural configurations during movement. Consequently, accurately estimating the robot’s shape is crucial to generating the substitute forward kinematics, thereby enabling the application of an InEKF state estimation framework as employed in legged robots.}


For shape reconstruction, which involves estimating the endcap positions within the body frame using onboard sensors, various approaches have been investigated. Elastic stretch sensors, as demonstrated in~\cite{tietz2013tetraspine, lu20226n}, are employed to measure the distance between two cable-connected endcaps. In~\cite{vespignani2018design}, motor encoders provide accurate cable length measurements for the nylon cables connecting the endcaps. IMUs installed on the structure can estimate the robot's orientation~\cite{huang2022live}. Li et al.~\cite{li2021shape} proposed a new shape recognition method using elastic cable length sensors and recurrent neural network (RNN), although this approach overlooks the tensegrity robot's geometrical properties. Booth et al.~\cite{booth2021surface} developed a sensor skin with conductive capacitive strain sensors to measure the distance between adjacent endcaps and used optimization to determine the endcap positions within the local reference frame.

For pose estimation, Moldagalieva et al.~\cite{moldagalieva2019computer} proposed a vision-based method for arm-like tensegrity robots using fiducial markers, assuming that the markers remain visible to overhead cameras.   
Previous work~\cite{caluwaerts2016state} combined proprioceptive and exteroceptive sensors, using an Unscented Kalman Filter (UKF) to estimate rods poses by fusing data from IMU and (Ultra-Wideband) UWB positioning sensor placed at the endcaps of the rods. More recently, Lu et al.~\cite{lu20226n} track the 6-DoF pose of each rod using overhead RGBD images combined with cable length measurements from onboard stretch sensors. Their method uses the correspondence between depth and RGB images to estimate rod poses during the transition step. In the correction step, joint-constrained optimization is performed with rod pose estimates and cable length measurements to refine estimates of the global rod poses.


In summary, while some progress has been made in shape and pose state estimation for tensegrity robots, no fully proprioceptive state estimator has been developed for tensegrity robots, which is the focus of this work. 


\section{Problem Formulation}

\subsection{Robot definition}
    \begin{figure}[t]
        \centering
        \includegraphics[width=0.9\linewidth]{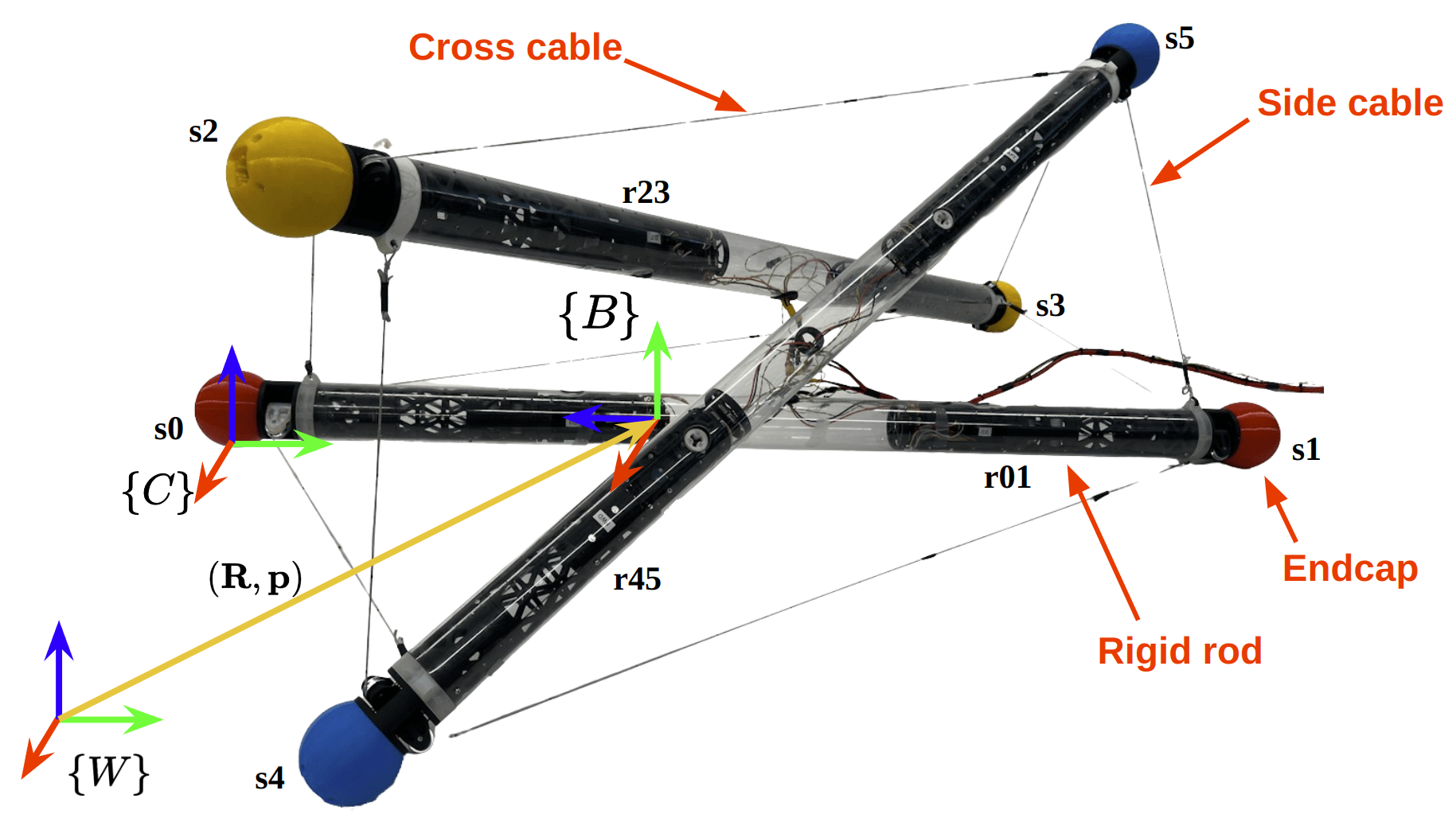}
        \caption{
        Tensegrity robot structure and coordination frames. Orientation $\mathbf{R}_t$ and position $\mathbf{p}_t$ of the robot are represented with respect to the world frame~($\mathbf{W}$). The IMU measurements $\mathbf{a}_t$, $\mathbf{\omega}_t$ are in the IMU frame, which is aligned with the Body frame~($\mathbf{B}$).
        }
        \label{fig:robot-label}
    \end{figure}

    \begin{figure*}[t]
        \centering
        \includegraphics[width=0.9\linewidth]{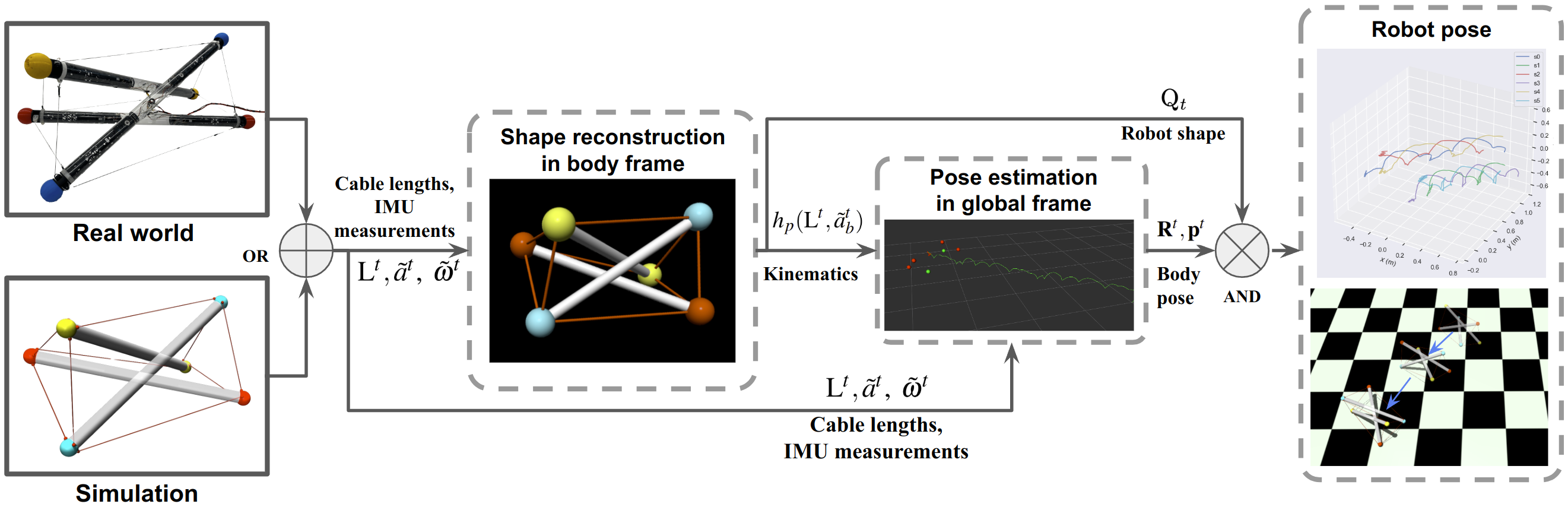}
        \caption{Tensegrity robot state estimation framework. Firstly, the real-world or simulated IMU and cable length sensors are input into an optimization-based robot shape reconstruction algorithm, as discussed in Sec.~\ref{sec: optimization}. The reconstructed shape provides the positions of the robot's endcaps in the body frame. Next, the computed kinematics, based on the contact points between the endcaps and ground, are utilized within a contact-aided Invariant EKF to estimate the robot pose. Finally, the global endcap positions are computed by transforming the reconstructed shape into the body pose within a global frame.
        }
        \label{fig:state-estimation-framewrok}
        \vspace{-4mm}
    \end{figure*}

    As shown in Fig.~\ref{fig:robot-label}, the 3-bar triangular prism tensegrity robot consists of three identical rigid rods with endcaps on each end. The endcaps on the same side denoted as $\{s_0, s_2, s_4\}$ and $\{s_1, s_3, s_5\}$ in Fig.~\ref{fig:robot-label}---are interconnected by three length controlled side cables. The endcaps on one end of each rod are also connected to the opposite endcaps of adjacent rods by three cross cables arranged in a counterclockwise configuration, thereby emulating the behavior of elastic springs. 
    This shape-morphing capability is exploited to facilitate rolling locomotion by shifting the robot's center of mass. 
    
    Unlike other robots, the tensegrity robot is rotationally symmetric, complicating the definition of a fixed robot base---essential for establishing the mathematical representation and kinematics of the robots. To address this, we designate the rod $r_{01}$ in Fig.~\ref{fig:robot-label} as the base rod. 
    The origin of the robot's body frame is defined at the position of the IMU, installed near the center of this base rod, denoted as $c_{01}$, with an offset of $d_\text{offset}$ towards $s_0$. The z-axis is aligned along the axial direction of the base rod. This specific frame definition is critical for developing the kinematic model and accurately describing the robot's motion.

\subsection{Problem definition}

    This work aims to track the \textbf{pose} of the 3-bar tensegrity robot and the \textbf{robot shape} in the body frame. i.e.,
    
    \begin{itemize}
        \item Tensegrity robot body pose in world frame: $\mathbf{x_t} \in \mathrm{SE(3)}$.
        \item Endcap positions in body frame at time $t$: $\mathrm{Q}_{t} = \left\{ \mathbf{q}^i_{t} | i \in \{0,1,2,3,4,5\} \right\} \subset \mathbb{R}^3$.
    \end{itemize}

    Given the sensor measurements and priors: 
    \begin{itemize}
        \item IMU acceleration and angular velocity measurements: $\tilde{a}_t$, $\tilde{\omega}_t$.
        \item Cable length measurements $\mathrm{L}_t = \left\{ \tilde{l}^{ij}_t | (i,j) \in \mathbb{E} \right\}$, where $i, j$ are the endcap ids. Here, $\mathbb{E}$ represents a set of pairs corresponding to the endcap ids connected by the actuated cables. Specifically,
        \begin{equation}
            \begin{aligned}
            \mathbb{E} = \{&(0,4), (0,2), (2,4), 
                           (1,5), (1,3), (3,5), \\
                           &(1,4), (0,3), (2,5) \} .
            \end{aligned}
        \end{equation}
        \item Endcap positions in body frame at $t-1$ time stamp:  $\mathrm{Q}_{t-1} = \left\{ \mathbf{q}^i_{t-1} | i \in \{0,1,2,3,4,5\} \right\} \subset \mathbb{R}^3$.
        \item $L_{\text{rod}}$ is a constant that represents the rigid bar's length.
    \end{itemize}

\section{Methodology}
\label{sec: methods}

The proprioceptive state estimation for the tensegrity robot inherits the legged robot state estimation frameworks in \cite{hartley2020contact, lin2023proprioceptive}, while the legged kinematic chain is substituted by the kinematics from the contact points to the robot base via tensegrity robot shape reconstruction in body frame. The framework overview is illustrated in Fig.~\ref{fig:state-estimation-framewrok}. 

Our assumptions are: a) all nine cable lengths can be accurately measured for robot shape reconstruction, which can be achieved by using stretch sensors \cite{johnson2022sensor, lu20226n} or motor encoders \cite{vespignani2018design}; b) contact events between the robot endcap and ground can be reliably detected; and c) the endcap-ground contact is modeled as point contact and remains static during the single contact period.

\subsection{State Representation on Lie Group}
    In robot state estimations, we focus on the robot's orientation, velocity, and position relative to a fixed world frame. We denote the rotational matrix $\mathbf{R}_t \in \mathrm{SO(3)}$ to represent the robot orientation and $\mathbf{v}_t \in \mathbb{R}^3$ and $\mathbf{p}_t \in \mathbb{R}^3$ as the body velocity and position in the world frame.
    In InEKF framework\cite{hartley2020contact, yu2023fully}, the state is defined in direct isometries group $\mathbf{x_t} \in \mathrm{SE_{l+2}(3)}$~\cite{barrau2015ekf, barrau2015non, luo2020geometry, lin2023proprioceptive}
    \begin{equation}
        \mathbf{x_t} := \left[\begin{array}{cccccc}
                    \mathbf{R}_t & \mathbf{v}_t & \mathbf{p}_t & \mathbf{d}_t^1 & \cdots & \mathbf{d}_t^l \\
                    \mathbf{0}_{l+2,3} & & & \mathbf{I}_{l+2} & &
                        \end{array}\right],
    \end{equation}
    where $\mathbf{d}^l_t \in \mathbb{R}^3$ is added as augmented states for foot contact position in the world frame if valid static contact occurs.

\subsection{IMU Propagation}
    \label{sec: imu-propagation}

    The IMU measures acceleration $a_t \in \mathbb{R}^3$ and angular velocity $\omega_t \in \mathbb{R}^3$. We model the IMU measurements as being corrupted by additive multivariate Gaussian noise \blue{ with biases denoted as $b_t^a, b_t^g \in \mathbb{R}^3$}:
    \begin{equation}
        \begin{aligned}
        \tilde{a}_t     & = a_t + b_t^a + w_t^a,      \quad w_t^a \sim \mathcal{N} \left(\mathbf{0}_{3,1}, \Sigma_a \right), \\
        \tilde{\omega}_t & = \omega_t + b_t^g + w_t^g, \quad w_t^g \sim \mathcal{N} \left(\mathbf{0}_{3,1}, \Sigma_g \right).
        \end{aligned}
    \end{equation}
    We assume that the endcap contact position in the world frame remains static throughout the contact period. However, to compensate for slippage and other uncertainties, we model the contact velocity as Gaussian white noise $w_t^d \sim \mathcal{N}(0, \Sigma^d)$. The unbiased continuous system dynamics can then be expressed as:
    \begin{subequations}
        \begin{align}
            \frac{d}{dt}{\mathbf{R}_t}   & = \mathbf{R}_t(\tilde{\omega}_t - w_t^g)_\times, \quad 
            \frac{d}{dt}{\mathbf{v}_t}    = \mathbf{R}_t(\tilde{a}_t - w_t^a) + \mathbf{g}, \\
            \frac{d}{dt}{\mathbf{p}_t}   & = \mathbf{v}_t, \quad 
            \frac{d}{dt}{\mathbf{d}_t}    = \mathbf{R}_t h_R(\mathrm{L}_t, \tilde a_t) (-w_t^d), \label{eqn: contact-dynamics}
        \end{align}
    \end{subequations}  
    where $(\cdot)_\times$ denotes the 3x3 skew-symmetric matrix, $\mathbf{g}$ represents the gravity vector, $\mathrm{L}_t$ is the cable length measurements, and $h_R(\mathrm{L}_t, \tilde a_t)$ represents the orientation of the contact frame from the shape optimization, which we will discuss in Sec.~\ref{sec: optimization}. We follow the method in \cite{hartley2020contact} to account for the system's biases.
    

    Rewriting into the matrix form, we have:
    \begin{equation}
    \small 
        \label{eqn: contact-imu-dynamics}
        \begin{aligned}
            &\frac{d}{dt}\mathbf{x}_t = 
            \left[ 
                \begin{array}{cccc}
                     \mathbf{R}_t(\tilde \omega_t)_\times & \mathbf{R}_t \tilde a_t + \mathbf{g} & \mathbf{v_t} & \mathbf{0}_{3,1}\\
                     \mathbf{0}_{1,3} & 0 & 0 & 0 \\
                     \mathbf{0}_{1,3} & 0 & 0 & 0 \\
                     \mathbf{0}_{1,3} & 0 & 0 & 0
                \end{array}
            \right] \\
            - &
            \left[
                \begin{array}{cccc}
                    \mathbf{R}_t & \mathbf{v}_t & \mathbf{p}_t & \mathbf{d}_t \\
                    \mathbf{0}_{1,3} & 1 & 0 & 0 \\
                    \mathbf{0}_{1,3} & 0 & 1 & 0 \\
                    \mathbf{0}_{1,3} & 0 & 0 & 1
                \end{array}
            \right]
            \left[
                \begin{array}{cccc}
                    (w_t^g)_\times & w^a_t & 0& h_R(\mathrm{L}_t, \tilde a^b_t) w_t^d \\
                    \mathbf{0}_{1,3} & 0 & 0 & 0 \\
                    \mathbf{0}_{1,3} & 0 & 0 & 0 \\
                    \mathbf{0}_{1,3} & 0 & 0 & 0
                \end{array}
            \right] \\
            & := f_{u_t} (\mathbf{x}_t) - \mathbf{x}_t {\mathbf{w}_t}^\wedge,
        \end{aligned}
    \end{equation}
    where, $\mathbf{w}_t := \mathrm{vec}(w^g_t, w^a_t, \mathbf{0}_{3,1}, h_R(\mathrm{L}_t, \tilde a_t) w_t^d )$ and $u_t := [(\tilde \omega_t)^\transpose, (\tilde a_t)^\transpose]^\transpose \in \mathbb{R}^6$. We can verify that the dynamic function $f_{u_t}(\cdot)$ satisfies the group affine condition, and the right invariant error is defined as: 
        $\eta_t^r=\bar{X}_t X_t^{-1}=\left(\bar{X}_t L\right)\left(X_t L\right)^{-1}$,
    where, $\bar{X_t}$ and $X_t$ represent two trajectories in the group $\mathcal{G}$, and $L \in \mathcal{G}$ is an arbitrary element of the group.

    Then, the right-invariant error dynamics\cite{hartley2020contact} is: 
    \begin{equation}
        \begin{aligned}
            \frac{d}{dt}\eta_t^r &= f_{u_t}(\eta^r_t) - \eta^r_t f_{u_t}(I) + (X_t \omega_t^\wedge X_t^{-1})\eta_t^r \\
            & :=g_{u_t}(\eta_t^r) + \omega_t^\wedge \eta^r_t
        \end{aligned}
    \end{equation}

    Then, from \cite{barrau2015non}, we can obtain the right-invariant state-independent linearized error dynamics by deriving $A_t^r$ from $g_{u_t}(\eta_t) = g_{u_t} (\exp(\xi_t)):=(A_t \xi_t)^\wedge + \mathcal{O}(\|\xi_t\|^2)$:    
   \begin{equation}
        \frac{d}{d t} \xi_t^r=A_t^r \xi_t^r-\operatorname{Ad}_{\bar{X}_t} w_t, 
        \quad A_t^r=\left[\begin{array}{cccc}
        0 & 0 & 0 & 0 \\
        (\mathbf{g})_{\times} & 0 & 0 & 0 \\
        0 & \mathbf{I} & 0 & 0 \\
        0 & 0 & 0 & 0
        \end{array}\right]
    \end{equation}

    In summary, with $A_t^r$ in, the state estimate $\bar{\mathbf{x}}_t$ is propagate through the system dynamics $f_{u_t}(\cdot)$. And the covariance matrix $P_t^r$ is computed using the Riccati equation:
    \begin{equation}
        \begin{aligned}
        \frac{d}{d t} \bar{\mathbf{x}}_t & =f_{u_t}\left(\bar{\mathbf{x}}_t\right) \\
        \frac{d}{d t} P_t^r & =A_t^r P_t^r+P_t^r A_t^{r \top}+\operatorname{Ad}_{\bar{\mathbf{x}}_t} Cov(w_t) \operatorname{Ad}_{\bar{\mathbf{x}}_t}^\transpose
        \end{aligned}
    \end{equation}

\subsection{Forward Kinematics Correction}
    Endcap contacts provide additional constraints for estimating the tensegrity robot's state by assuming contact velocity is zero in the world frame.

    The contact state $C$ is defined as a binary vector $\mathbf{c}_t = [c_0, c_1, c_2, c_3, c_4, c_5]$, where $c_i \in \{0, 1\}$ for each endcap $s_i$. The contact state can be estimated using a robot shape and IMU reading or an external contact sensor. 

    \subsubsection{Contact State Augmentation}
        Once the contact event $C$ is detected, we append the contact state into the state variable using the optimized robot shape information in Sec.\ref{sec: optimization}, which is: 
                $\bar{\mathbf{d}^i_t} = \mathbf{p}_t + \mathbf{R}_t h_p(\mathrm{L}_t, \tilde a_t)$,
        where $h_p(\cdot)$ is the optimization function that takes the cable length measurements and IMU accelerometer readings to calculate the ground-contact endcap position in the body frame. 
        The corresponding covariance can be augmented using:
        \begin{equation}
            \begin{aligned}
            P_{t_k}^{\text {new }} & =F_{t} P_{t} F_{t}^\transpose+G_{t} \operatorname{Cov}\left(w_{t}^a\right) G_{t_k}^\transpose, \\
            F_{t_k} & =\left[\begin{array}{lll}
            \mathbf{I} & 0 & 0 \\
            0 & \mathbf{I} & 0 \\
            0 & 0 & \mathbf{I} \\
            0 & 0 & \mathbf{I}
            \end{array}\right], \quad G_{t_k}=\left[\begin{array}{c}
            0 \\
            0 \\
            0 \\
            \bar{R}_{t_k} J_p\left( \mathrm{L}_t, \tilde a_t \right)
            \end{array}\right],
            \end{aligned}
        \end{equation}
        where $J_p\left( \cdot \right)$ is the Jacobian of $h_p(\cdot)$. For the tensegrity robot, the $h_p(\cdot)$ is formulated as an optimization problem, and we cannot derive the Jacobian analytically. Instead, we treat it as a tunable parameter and empirically determine the covariance value.
        The augmented contact state $\mathbf{d}_t^i$ remains in the robot state $\mathbf{x}_t$ throughout the single contact phase. When contact is terminated, the contact state is marginalized out of the robot state by:
        \begin{equation}
        \small 
            \bar{\mathbf{x}}_{t_k}^{\text {new }}=M \bar{\mathbf{x}}_{t_k}, \bar{P}_{t_k}^{\text {new }}=M \bar{P}_{t_k} M^\transpose, M=\left[\begin{array}{cccc}
            I & 0 & 0 & 0 \\
            0 & I & 0 & 0 \\
            0 & 0 & I & 0
            \end{array}\right]
        \end{equation}

    \subsubsection{Correction Model}
        Once the endcap enters the contact phase, the state can be corrected using the right-invariant correction model:
        \begin{equation}
        \small 
            \begin{aligned}
            Y_{t_k} &= X_{t_k}^{-1} b + V_{t_k} \\
            \begin{bmatrix}
                h_p( \mathrm{L}_t, \tilde a_t ) \\
                0\\
                1\\
                -1
            \end{bmatrix}
            &= \begin{bmatrix}
                R_{t_k}^\transpose & -R_{t_k}^\transpose v_{t_k} & -R_{t_k}^\transpose p_{t_k} & -R_{t_k}^\transpose d_{t_k} \\
                0 & 1 & 0 & 0 \\
                0 & 0 & 1 & 0 \\
                0 & 0 & 0 & 1
            \end{bmatrix}
            \begin{bmatrix}
                0\\
                0 \\
                1 \\
                -1
            \end{bmatrix}\\
            &+\begin{bmatrix}
                J_p( \mathrm{L}_t, \tilde a_t ) w^a_{t_k}\\
                0\\
                0 \\
                \blue{0}
            \end{bmatrix},
            \end{aligned}
        \end{equation}
        The linearized measurement matrix $H$ and the noise matrix $N_k$ are:
        \begin{equation}
            \begin{aligned}
                H &= \begin{bmatrix}
                    0 & 0 & -I & I
                \end{bmatrix},\\
                \bar N_k &= \bar{R}_{t_k} J_p( \mathrm{L}_t, \tilde a_t ) \operatorname{Cov}(w^a_{t_k}) J_p( \mathrm{L}_t, \tilde a_t )^\transpose \bar{R}_{t_k}^\transpose .
            \end{aligned}
        \end{equation}
        Finally, the RI-EKF contact correction can be calculated using
        \begin{equation}
            \begin{aligned} 
            \label{eqn:riekf-correction}
                   \bar{X}_{t_k}^+ &= \exp \left( L_{t_k} \left( \bar{X}_{t_k} Y_{t_k} - b \right) \right) \bar{X}_{t_k}, \\
                P_{t_k}^{r+} &= (I - L_{t_k} H) P_{t_k}^r (I - L_{t_k} H)^\transpose + L_{t_k} \bar{N}_k L_{t_k}^\transpose,
            \end{aligned}
        \end{equation}
        where the Kalman gain $L_{t_k}$ and innovation covariance $S_{t_k}$ are
        \begin{equation}
            L_{t_k} = P_{t_k}^r H^\transpose S^{-1}, \quad S = H P_{t_k}^r H^\transpose + \bar{N}_k .
        \end{equation}

\section{Shape Reconstruction}
 \subsection{Optimization Formulation}
    \label{sec: optimization}
    To solve for $h(\mathrm{L}_t, \tilde a_t)$ as required in Eqn.~\eqref{eqn: contact-imu-dynamics} in Sec.~\ref{sec: imu-propagation}, we developed a constrained optimization algorithm that utilizes cable length measurements and IMU acceleration data to estimate the position of endcaps in the body frame.
    
     The intuition behind this optimization is that the distances between the estimated endcap positions should closely match the measured cable length. Additionally, the endcap positions and rods must satisfy a set of constraints to ensure the robot's shape, i.e., the endcap positions in the body frame, are valid. 
     The goal of the optimization problem in Eqn.~\eqref{eqn: optimization} is to find the optimal set of endcap positions $\mathrm{Q}_t^* = \left\{ \mathbf{q}^i_t \in \mathbb{R}^3 | i \in \{0, 1, 2, 3, 4, 5\} \right\}.$

     The variable $\mathbf{q}_t^{i,j}$, where $j \in {1, 2, 3}$, denotes the $j$-th component of the endcap position vector $\mathbf{q}_t^i$. Furthermore, $\mathbf{c}_t^{ij}$ represents the coordinates of the center of the rigid bar $r_{ij}$.
    \begin{subequations}
        \label{eqn: optimization}
        \begin{align}
        & \mathrm{minimize} \sum_{(i, j) \in \mathbb{E}} \left(\left\|\mathbf{q}^i_t-\mathbf{q}^j_t\right\|-\tilde{l_t^{ij}}\right)^2 \label{eqn: optimization obj}\\
        \mathrm{s.t.}  \quad & \mathbf{q}^0_t = [0, 0, L_{\text{rod}}/2-d_\text{offset}]^\transpose  \label{eqn: optimization-constraints-endcap0} \\
                                 & \mathbf{q}^1_t = [0, 0, -L_{\text{rod}}/2-d_\text{offset}]^\transpose \label{eqn: optimization-constraints-endcap1} \\
                                 & \mathbf{q}_t^{2, 3} > 0, \mathbf{q}_t^{3, 3} < 0 \label{eqn: optimization-constraints-endcap23} \\
                                 & \mathbf{q}_t^{4, 3} > 0, \mathbf{q}_t^{5, 3} < 0 \label{eqn: optimization-constraints-endcap45} \\
                                 & \|\mathbf{q}^0_t - \mathbf{q}^1_t \| = \|\mathbf{q}^2_t - \mathbf{q}^3_t \| = \|\mathbf{q}^4_t - \mathbf{q}^5_t \| = L_{\text{rod}} \label{eqn: optimization-constraints-between-endcaps} \\
                                 & \left< (\mathbf{c}^{23}_t-\mathbf{c}^{01}_t) \times (\mathbf{c}^{45}_t - \mathbf{c}^{23}_t) , \mathbf{q}^2_t - \mathbf{q}^3_t \right> > 0 \label{eqn: optimization-constraints-cross0} \\
                                 & \left< (\mathbf{c}^{45}_t-\mathbf{c}^{23}_t) \times (\mathbf{c}^{01}_t - \mathbf{c}^{45}_t), \mathbf{q}^4_t - \mathbf{q}^5_t \right> > 0 \label{eqn: optimization-constraints-cross1} \\
                                 & \left< (\mathbf{c}^{01}_t-\mathbf{c}^{45}_t) \times (\mathbf{c}^{23}_t - \mathbf{c}^{01}_t) , \mathbf{q}^1_t - \mathbf{q}^0_t \right> > 0 \label{eqn: optimization-constraints-cross2} \\
                                 & \left< \mathbf{q}_t^2 - \mathbf{q}_t^4, \mathbf{q}_t^5 - \mathbf{q}_t^1  \right> > 0 \label{eqn: optimization-constratins-twist0} \\
                                 & \left< \mathbf{q}_t^0 - \mathbf{q}_t^2, \mathbf{q}_t^3 - \mathbf{q}_t^5  \right> > 0 \label{eqn: optimization-constratins-twist1} \\
                                 & \left< \mathbf{q}_t^4 - \mathbf{q}_t^0, \mathbf{q}_t^1 - \mathbf{q}_t^3  \right> > 0 \label{eqn: optimization-constratins-twist2}
        \end{align}
    \end{subequations}
    \subsection{Optimization variable \& objectives}
        $\mathbf{q^i_t}$ represents the body frame position of endcap $s_i$ at time $t$. The set of positions to be optimized is $\mathrm{Q}_t = \left\{ \mathbf{q}^i_t \in \mathbb{R}^3 | i \in \{0, 1, 2, 3, 4, 5\} \right\}$. 
        The optimization variable $\mathbf{q}_t := {[{\mathbf{q}_t^0}^\transpose, {\mathbf{q}_t^1}^\transpose, {\mathbf{q}_t^2}^\transpose, {\mathbf{q}_t^3}^\transpose, {\mathbf{q}_t^4}^\transpose, {\mathbf{q}_t^5}^\transpose]}^\transpose$.

        Consequently, the objective function minimizes the error between the measured endcap distances and the estimated endcap distances is denoted as $\|\mathbf{q}^i_t-\mathbf{q}^j_t\|, (i, j) \in \mathbb{E}$, and the sensor reading of the nine actuation cables as $\tilde{l_t^{ij}}$. 
    
    \subsection{Optimization constraints}
        The equality and inequality constraints of the optimization problem are as follows: 
        
        \noindent \textbf{Endcap arrangement constraints:} Eqn.~\eqref{eqn: optimization-constraints-endcap0}-\eqref{eqn: optimization-constraints-endcap1} represent the fixed coordinates of $s_0$, $s_1$ in the body frame as shown in Fig.~\ref{fig:robot-label}. Eqn.~\eqref{eqn: optimization-constraints-endcap23}-\eqref{eqn: optimization-constraints-endcap45} represent the third element (z-axis) of $\mathbf{q}^2, \mathbf{q}^4$ and $\mathbf{q}^3, \mathbf{q}^5$, which are $>0$ and $<0$ , respectively, i.e., $s_0, s_2, s_3$ are in the upper-half-plane, while $s_1, s_3, s_5$ are in the lower-half-plane of the body frame.
        
        \noindent \textbf{Rod length constraints:} Eqn.~\eqref{eqn: optimization-constraints-between-endcaps} defines constraints between endcaps. Since the rods in tensegrity robots are rigid, the distance between any two endcaps connected by a rod must remain invariant, equaling $L_{\text{rod}}$ at all times.

\begin{figure}[t]
            \centering
            \includegraphics[width=0.75\columnwidth]{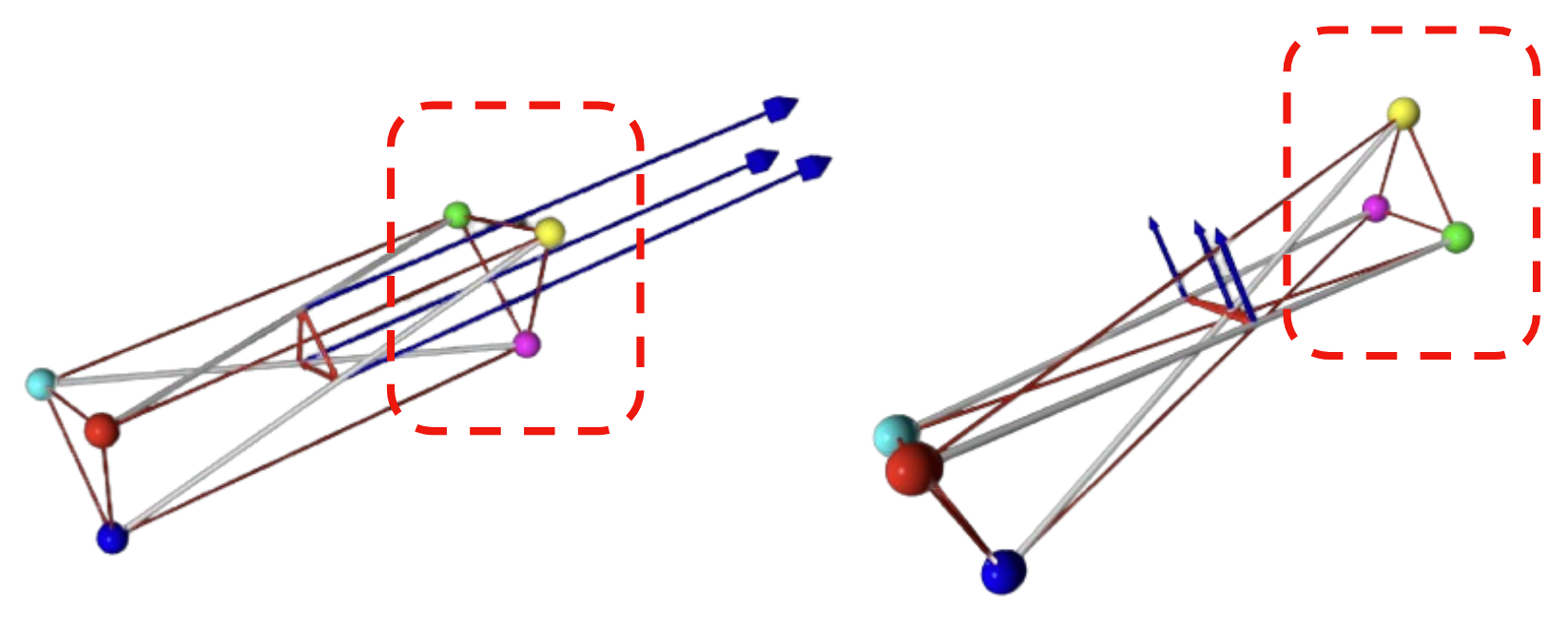}
            \caption{Illustration of geometric constraints, red arrows represent the vectors chained to each rod center, blue arrows represent their cross-product vectors. 
            In the left figure, the geometric constraints are satisfied, with the cross-product vector pointing along the robot's axial direction to the right. Conversely, the right figure demonstrates a switch to the opposite direction. The endcap colors in the circled area follow yellow-green-pink in the left figure and yellow-pink-green in the right, demonstrating the difference in the robot configuration.
            }
            \label{fig:geometric constraints}
            \vspace{-6mm}
        \end{figure}

        \noindent \textbf{Geometric Constraints:} Eqn.~\eqref{eqn: optimization-constraints-cross0}-\eqref{eqn: optimization-constraints-cross2} are geometric constraints to prevent the rods from crossing each other over time. $\mathbf{c}_t^{23} - \mathbf{c}_t^{01}$ represents the vector from the center point of the rod $r_{01}$, to the center of rod $r_{23}$. The cross-product of the chained center vectors should point along the axial axis, as shown in Fig.~\ref{fig:geometric constraints}, rather than in the opposite or radial direction. Thus, the dot product of the cross-product vector and the axial vector remains positive.
        
\begin{figure}[t]
                \centering
                \includegraphics[width=0.8\linewidth]{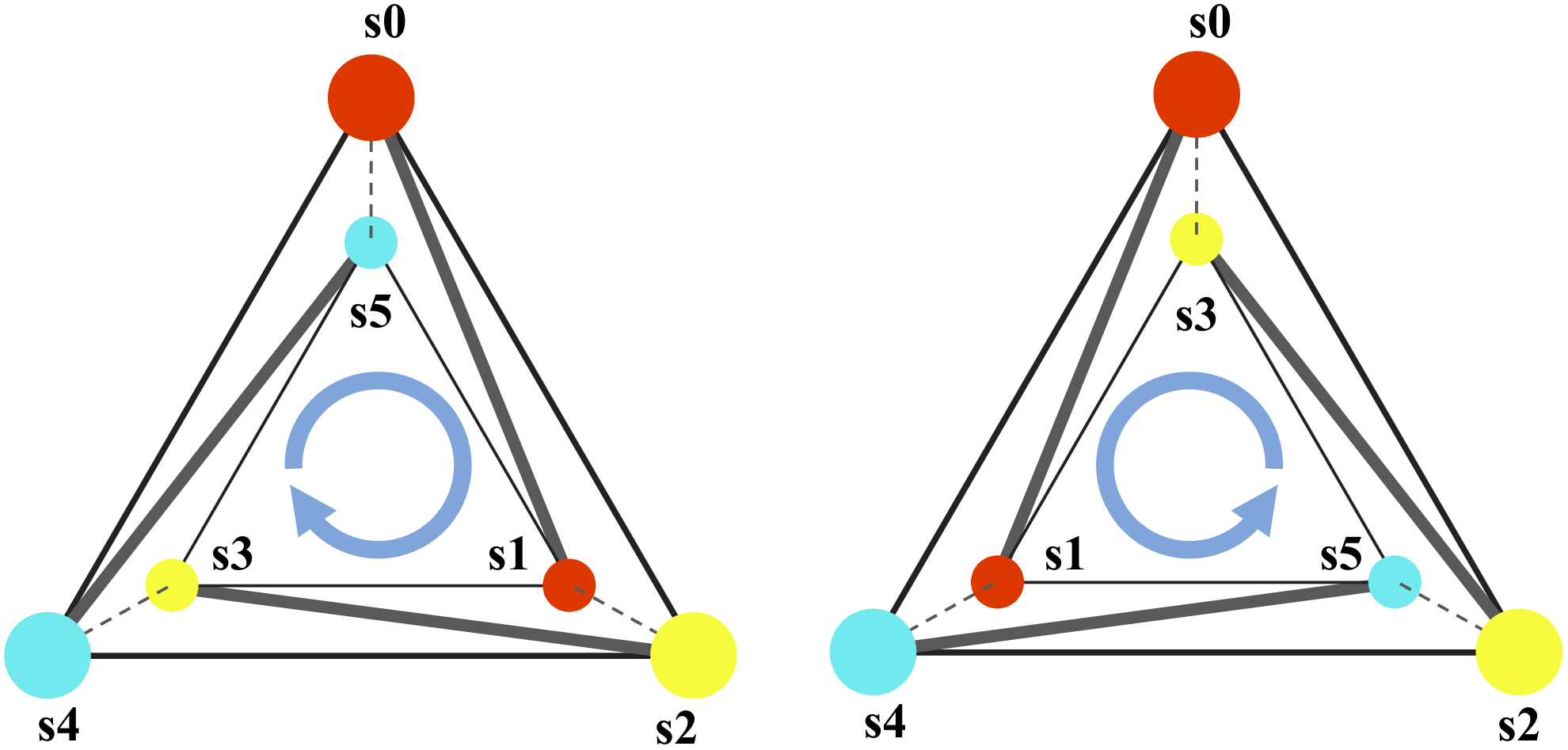}
                \caption{Illustration of 3-bar tensegrity robot chirality from side-view. The right configuration is twisting counter-clockwise, the same structure we used in simulation and real-world robots. }
                \label{fig:tensegrity-chirality}
                \vspace{-4mm}
            \end{figure}

        \noindent \textbf{Chirality Constraints:} Eqn.~\eqref{eqn: optimization-constratins-twist0}-\eqref{eqn: optimization-constratins-twist2} are chirality constraints. The tensegrity structure forms a twisted 3-prism, which can have two twist directions as shown in Fig.~\ref{fig:tensegrity-chirality}. The tensegrity robot in both the simulation and the real world follows the right configuration in Fig.~\ref{fig:tensegrity-chirality}, where the angles between two corresponding vectors (e.g., $\mathbf{q}^{24}, \mathbf{q}^{15}$) on the other side are acute, satisfying the constraints. 
        
\section{Experiments}
\subsection{Experiments Setup}
\label{sec: experiments-setup}
    We conducted experiments with the cable-actuated tensegrity robot\cite{mi2024design} in both simulation and real-world settings. The tensegrity robot consists of three bars, each measuring a length of 1.45m, 76mm in diameter, and weighing 4.0kg. 
    An IMU (VectorNav VN-100) is attached at the center of the bar with red endcaps, providing linear acceleration and angular velocity data at 200Hz. \blue{The inelastic cables are actuated by brushless DC motors with an AS5047P encoder, which operates at 100Hz. The motors are controlled to apply a minimum torque, ensuring cables remain under tension and preventing slackness\cite{mi2024design}. The cable lengths are determined from the encoder values with zero-slackness assumption. The distance between the endcaps is then calculated by adding the endcap radius to the measured cable lengths.}
    Due to the current hardware limitations, onboard contact sensing has not yet been implemented. Instead, we rely on the Vicon motion capture system (MoCAP) for contact vector $\mathbf{c}_t$. 
    In the MuJoCo simulation, the physical properties of the robot, including dimensions, mass, moment of inertia (MoI), and cable spring constants, as well as sensor biases and noises, are characterized to match the real-world tensegrity robot~\cite{mi2024design} (see Table.~\ref{tab:noise-param} for the parameters).

    \begin{table}[t]
        \centering
        \caption{Experiment Noise Parameters}
        \begin{tabular}{l|l}
            \hline
            Measurement Type & Noise STD \\
            \hline
            Linear acceleration & 0.043 $m/s^2$ \\
            Angular velocity    & 0.002 $rad/s$ \\
            Accelerometer bias &  0.001 $m/s^3$ \\
            Gyroscope bias     &  0.001 $rad/s^2$\\
            Contact linear velocity & 0.05 $m/s$ \\
            \hline
        \end{tabular}
        \label{tab:noise-param}
        \vspace{-3mm}
    \end{table}

    
\subsection{Simulation Shape Reconstruction Results}
\begin{figure}[t]
        \centering
        \includegraphics[width=0.95\linewidth]{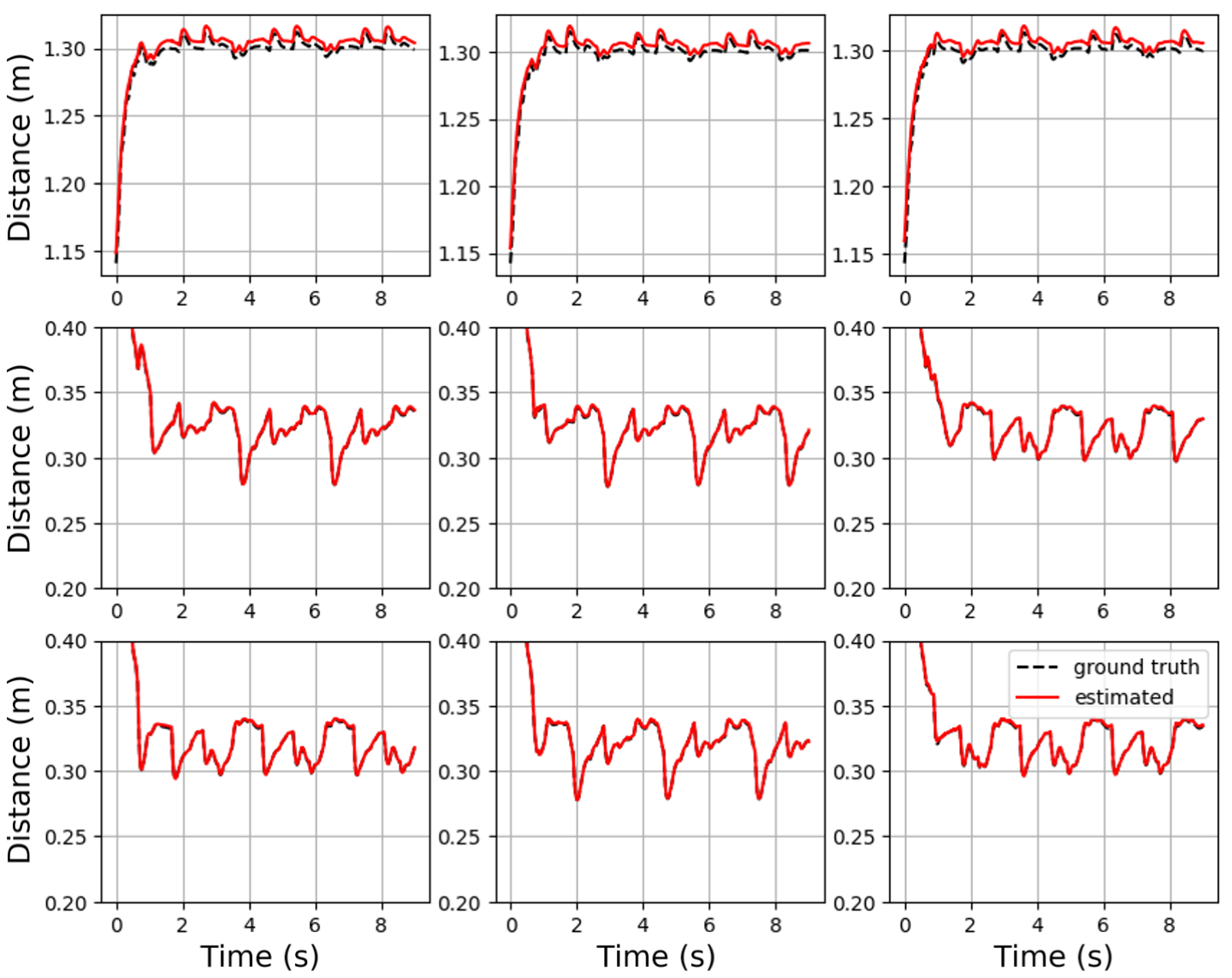}
        \caption{Plots of the optimized cable lengths (red) compared to the ground truth distance between endcaps (black).}
        \label{fig:result-tendon-length}
        \vspace{-6mm}
    \end{figure}

    We quantitatively evaluated shape reconstruction by comparing the distances between the estimated cable-connected endcap positions $\left\| \mathbf{q_t^i} - \mathbf{q_t^j} \right\|, (i,j) \in \mathrm{E}$ and the ground-truth cable length $l_t^{ij}$ from the simulation, as shown in Fig.~\ref{fig:result-tendon-length}. Without incorporating geometric constraints, significant shape reconstruction errors were observed due to the opposite chirality of the tensegrity structure, as illustrated in Fig.~\ref{fig:geometric constraints} (right).
    By incorporating geometric constraints, the RMSE was reduced to 4.63 cm, which is negligible given the robot's bar length of 1.45 m. 
    
\subsection{Simulation Pose Estimation Results}
    \label{sec: simulation-results}
    Our pose estimation method builds on the DRIFT\cite{lin2023proprioceptive} Invariant EKF framework.
    We conducted \blue{five} simulation experiments: \textit{move-forward}, \textit{move-backward}, \textit{right-turn}, \textit{left-turn}, and \blue{\textit{valley}}, each generating a different trajectory. At the beginning of each sequence, the robot remained stationary for a few seconds for sensor initialization, such as accelerometer bias calibration. The robot was controlled via motor-actuated cables to perform predefined maneuvers. In the simulation, contacts were detected for the contact kinematic correction step in the Right Invariant EKF. 
    IMU measurements, cable length measurements, and contact vectors were published using ROS in the MuJoCo simulation node, \blue{while the ground-truth poses are logged for results comparison.} The shape reconstruction node subscribed to these inputs and published the robot shape $\mathbf{Q}$, while the Invariant EKF pose estimator estimated the robot pose in real-time, recording both the estimated and ground-truth trajectories for later comparison. We employ evo \cite{grupp2017evo} to align the initial segment of the estimated trajectory with the ground truth trajectory, as the initial pose is unobservable.

\begin{figure}[t]
        \centering
        \includegraphics[width=0.9\linewidth]{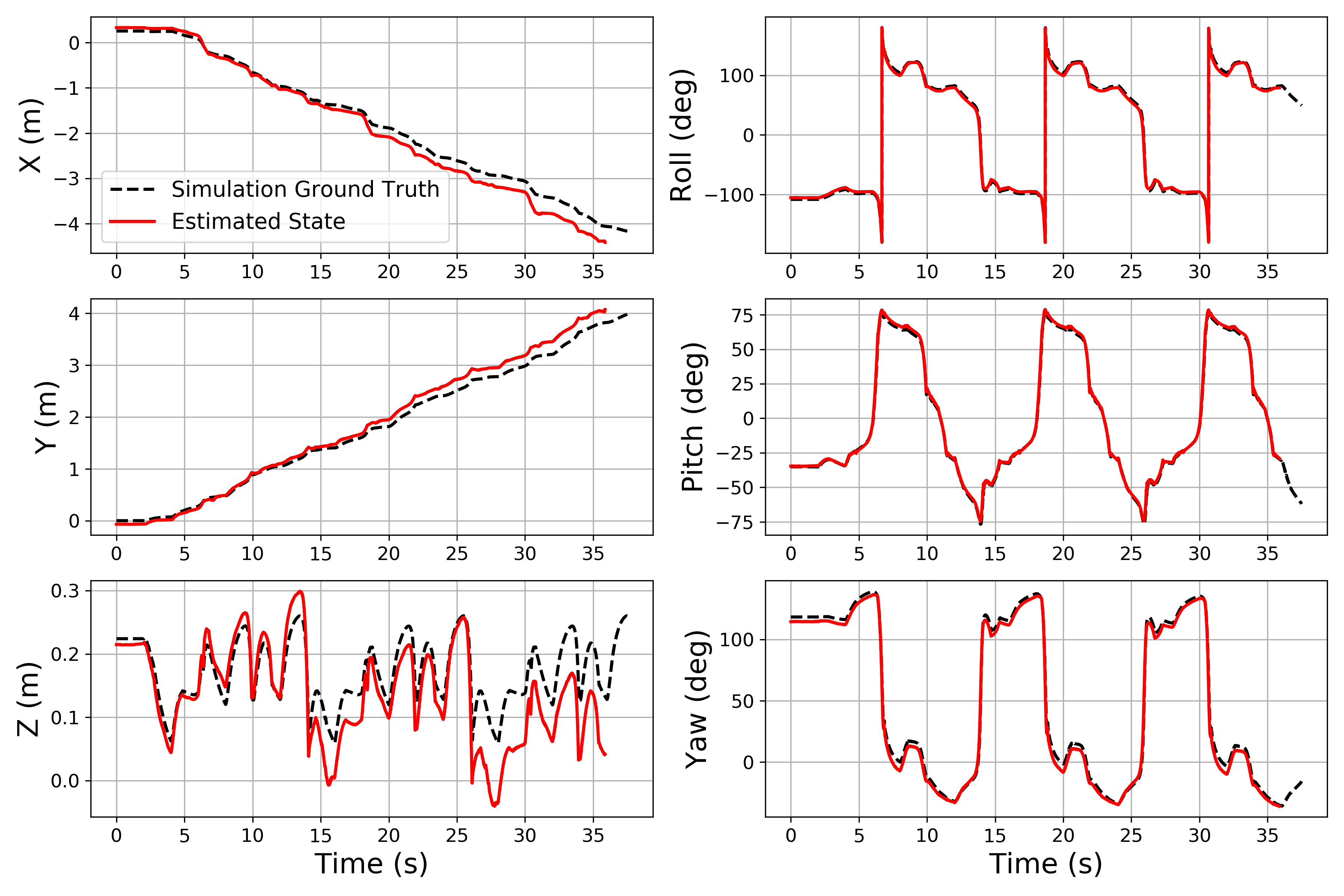}
        \caption{Estimated position and orientation compared with ground truth in \textit{forward} dataset.}
        \label{fig:result-sim-linear-pos-ori}
        \vspace{-3mm}
    \end{figure}

    Fig.~\ref{fig:result-sim-linear-pos-ori} presents the estimated position and orientation compared to the ground truth for the \textit{forward} movement dataset, demonstrating robust tracking of both position and orientation. The deviations in the X and Y positions suggest the estimator is well-suited for translational movements, although larger discrepancies were observed along the Z-axis. The orientation estimation captures the oscillatory motions and is generally aligned with the ground truth.
    
    \begin{figure}[t]
        \centering
        \includegraphics[width=0.9\linewidth]{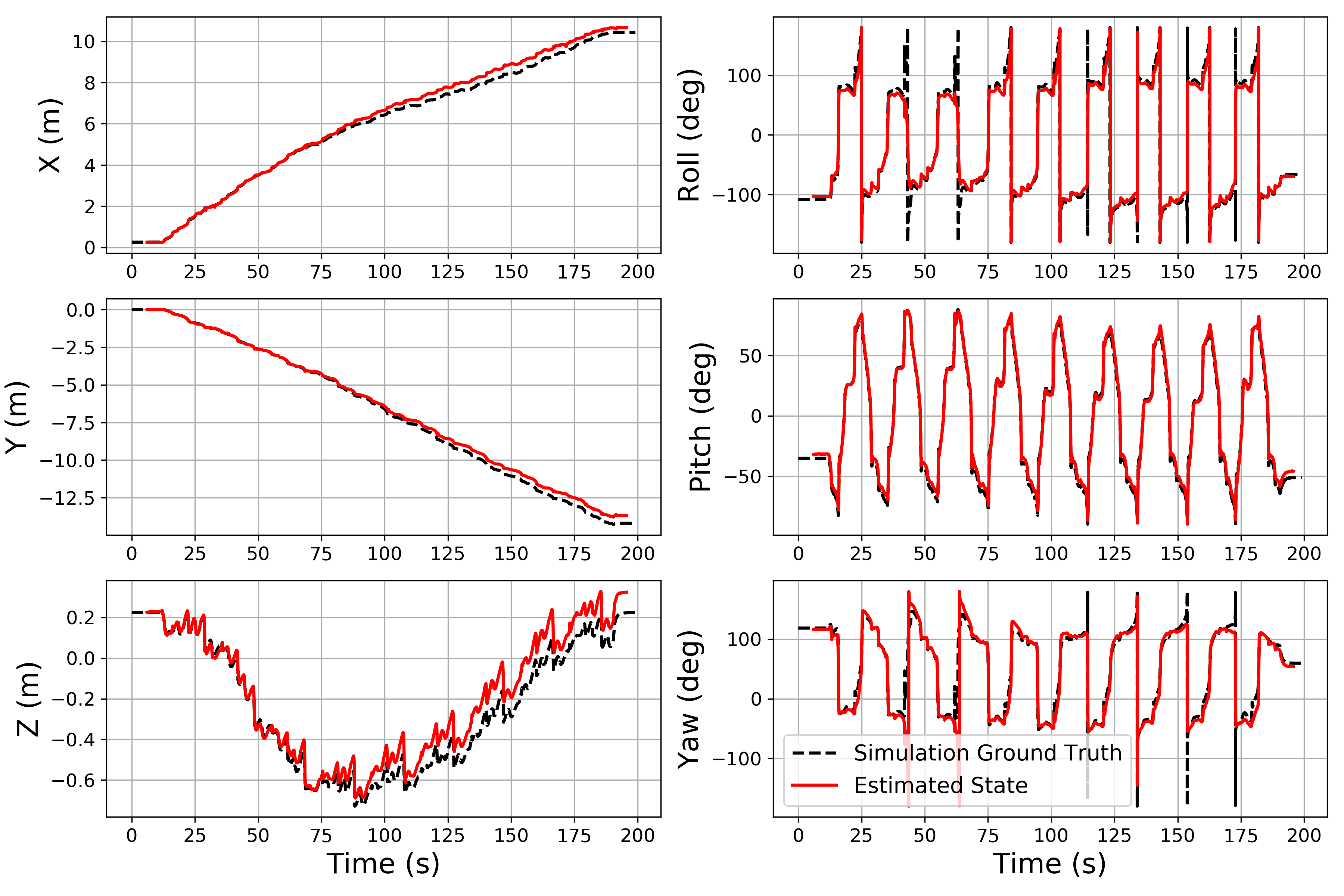}
        \caption{\blue{The estimated position and orientation are compared with ground truth in the \textit{valley} dataset, which features a 15-degree downhill followed by a 15-degree uphill scenario.}}
        \label{fig:result-sim-valley-pos-ori}
        \vspace{-6mm}
    \end{figure}

\begin{figure}[t]
        \centering
        \includegraphics[width=0.9\linewidth]{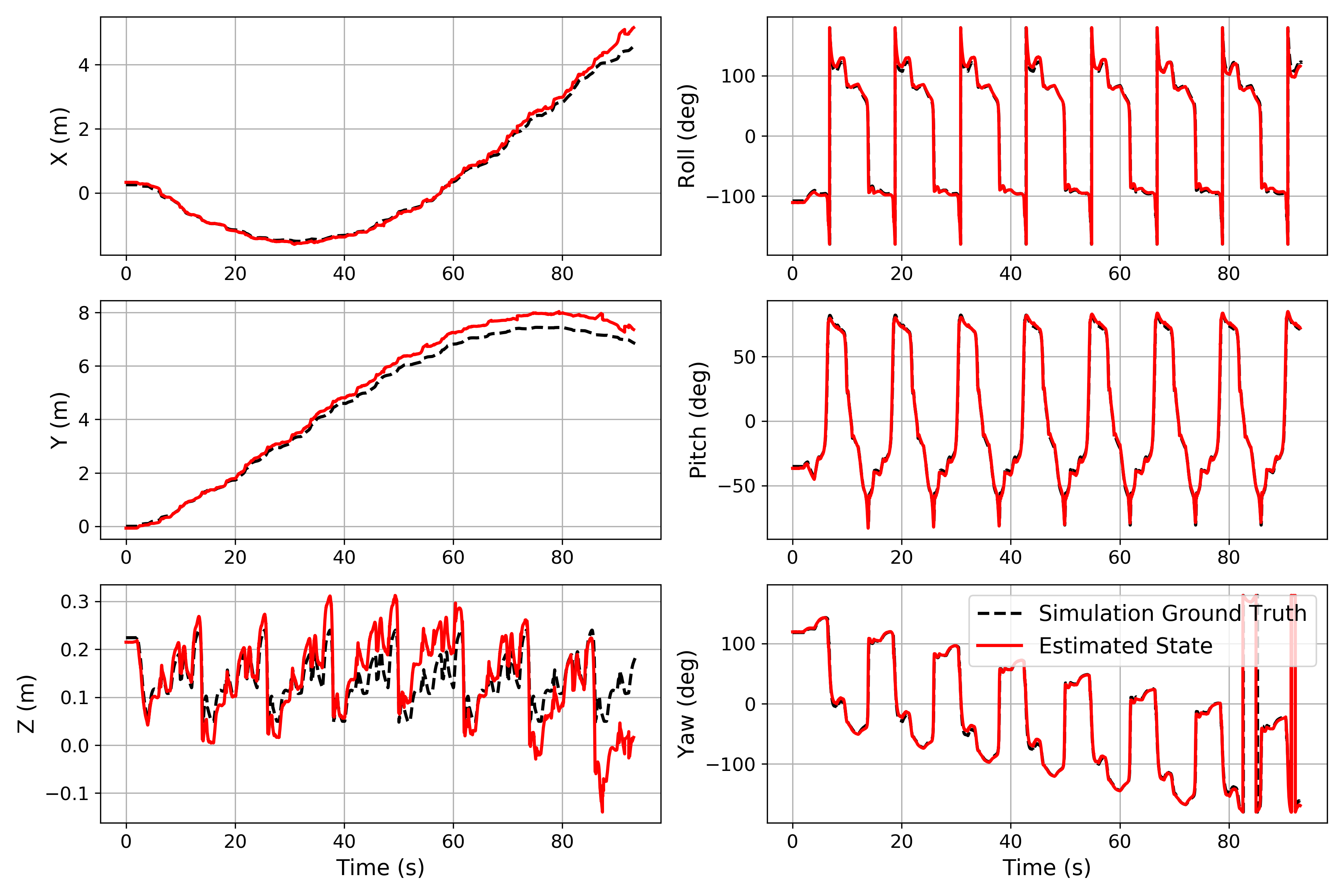}
        \caption{Estimated position and orientation compared with ground truth in \textit{right-turn} dataset.}
        \label{fig:result-sim-right-turn-pos-pri}
        \vspace{-3mm}
    \end{figure}

\begin{figure}[t]
        \centering
        \includegraphics[width=0.65\linewidth]{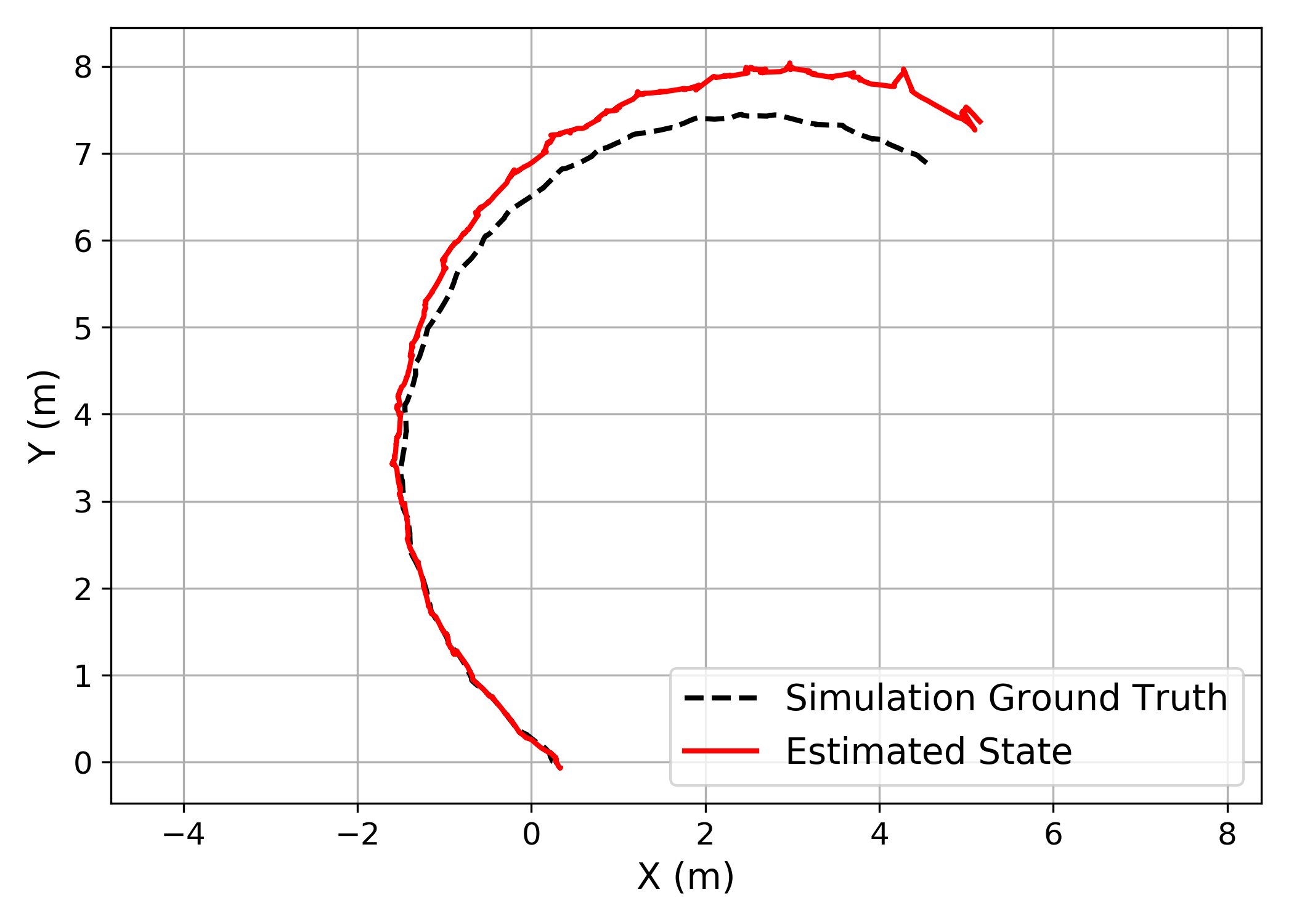}
        \caption{Topview of \textit{right-turn} dataset estimated trajectory compared with ground-truth.}
        \label{fig:result-sim-right-turn-traj}
        \vspace{-6mm}
    \end{figure}

    \blue{In the \textit{right-turn} and \textit{valley} datasets, shown in Fig.~\ref{fig:result-sim-right-turn-traj} and Fig.~\ref{fig:result-sim-valley-pos-ori}, the robot successfully navigated longer trajectories of 15.70 m and 22.19 m, respectively. Despite a final drift of 0.76 m over the 15.70-meter trajectory in the \textit{right-turn} dataset, the drift percentage remained low at 4.84\%. Similarly, the \textit{valley} dataset reported a low drift percentage of 4.15\%. As illustrated in Fig.~\ref{fig:result-sim-right-turn-pos-pri} and Fig.~\ref{fig:result-sim-valley-pos-ori}, the estimator demonstrated comparable qualitative performance to the \textit{forward} dataset, highlighting its consistent accuracy even during more complex maneuvers and on challenging terrain.}

    \begin{table}[t]
        \centering
        \caption{Simulation and Real-world Experiments Statistics in Meter and Degree. We report the RMSE of Relative Pose Error (RPE) in the unit of drift per meter.}
        \resizebox{\columnwidth}{!}{
            \begin{tabular}{l|l|l|l|l|l|l}
            \hline
                Metric                   & \textit{Forward} & \textit{Backward} & \textit{Left-turn} & \textit{Right-turn} & \textit{Forward-realworld} & \blue{valley} \\
            \hline
                Traj. Length             & 7.50             & 6.74              & 18.61              & 15.70               & 5.02                    & \blue{22.19 } \\
                Final Drift              & 0.3275           & 0.3386            & 0.8221             & 0.7610              & 0.1174                  & \blue{0.92}  \\
                Drift percentage         & 4.37\%           & 5.02\%            & 4.42\%             & 4.85\%              & 2.34\%                  & \blue{4.15\%}  \\
                RPE RMSE Trans.          & 0.0911           & 0.0936            & 0.1159             & 0.1022              & 0.1927                  & \blue{0.0735}  \\
                RPE RMSE Rot.            & 2.86             & 3.72              & 3.94               & 3.18                & 10.84                   & \blue{4.13}  \\
            \hline
            \end{tabular}
        }
        \label{tab:res-pose-error}
        \vspace{-2pt}
    \end{table}
    
\subsection{Real-world Pose Estimation Results}
\begin{figure}[htbp]
        \centering
        \includegraphics[width=0.7\linewidth]{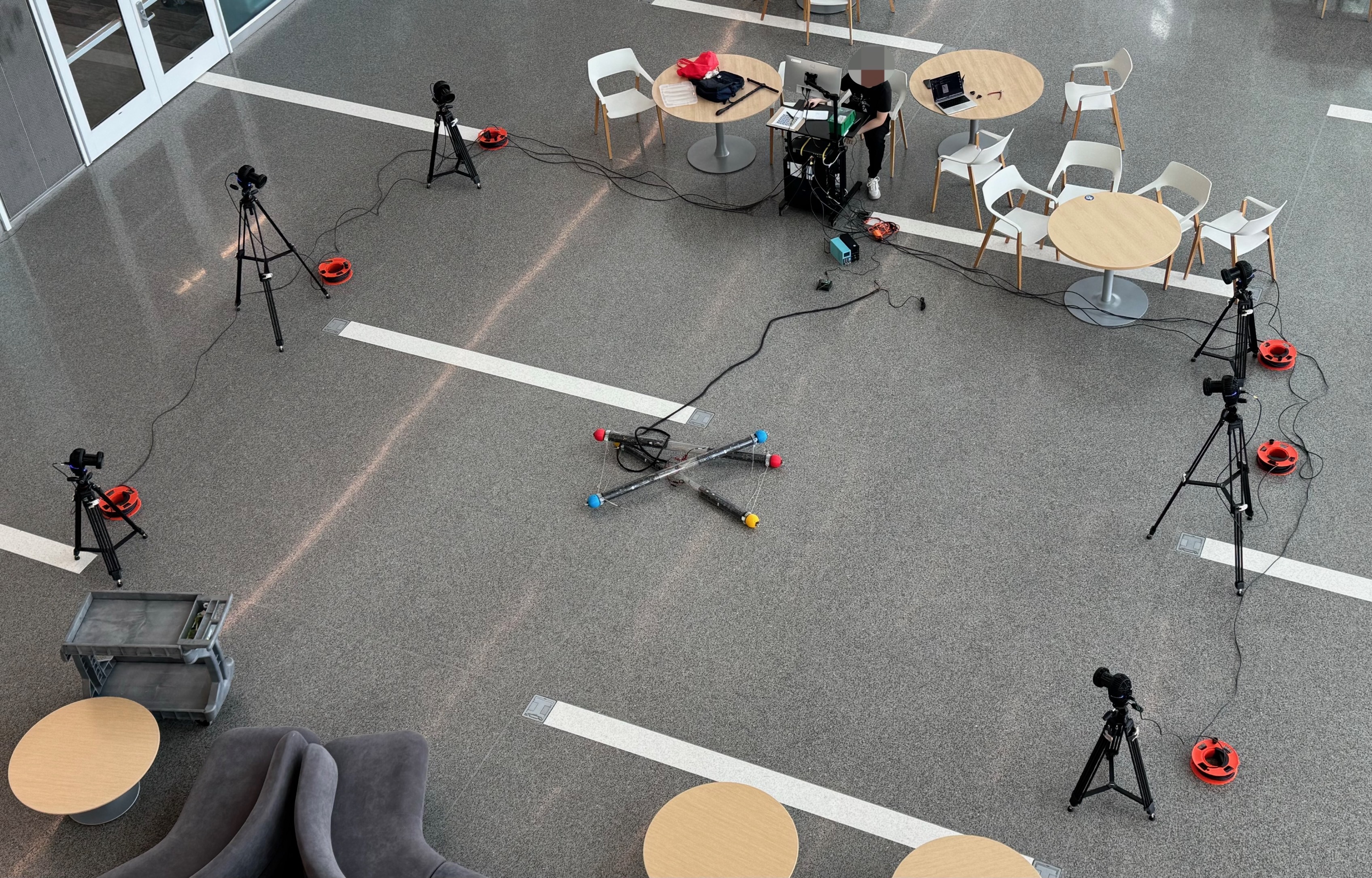}
        \caption{Experiments setup with motion capture system to acquire ground truth position and orientation data for the tensegrity robots.}
        \label{fig:result-test-field}
        \vspace{-6mm}
    \end{figure}

    We conducted analogous experiments in a real-world environment to those in the \textit{forward} simulation. Six Vicon motion capture cameras were arranged as shown in Fig.~\ref{fig:result-test-field}, with reflective markers affixed to the tips of the robot's endcaps. The MoCAP system tracked these markers to compute the ground-truth robot state and detect endcap-ground contacts, as detailed in Sec.~\ref{sec: experiments-setup}. The VN-100 IMU and motor controller board inside the tensegrity robot were interfaced with an Nvidia Jetson Orin Nano, with motion capture data synchronized via ROS. All collected data were archived for subsequent analysis.

\begin{figure}[t]
        \centering
        \includegraphics[width=0.9\linewidth]{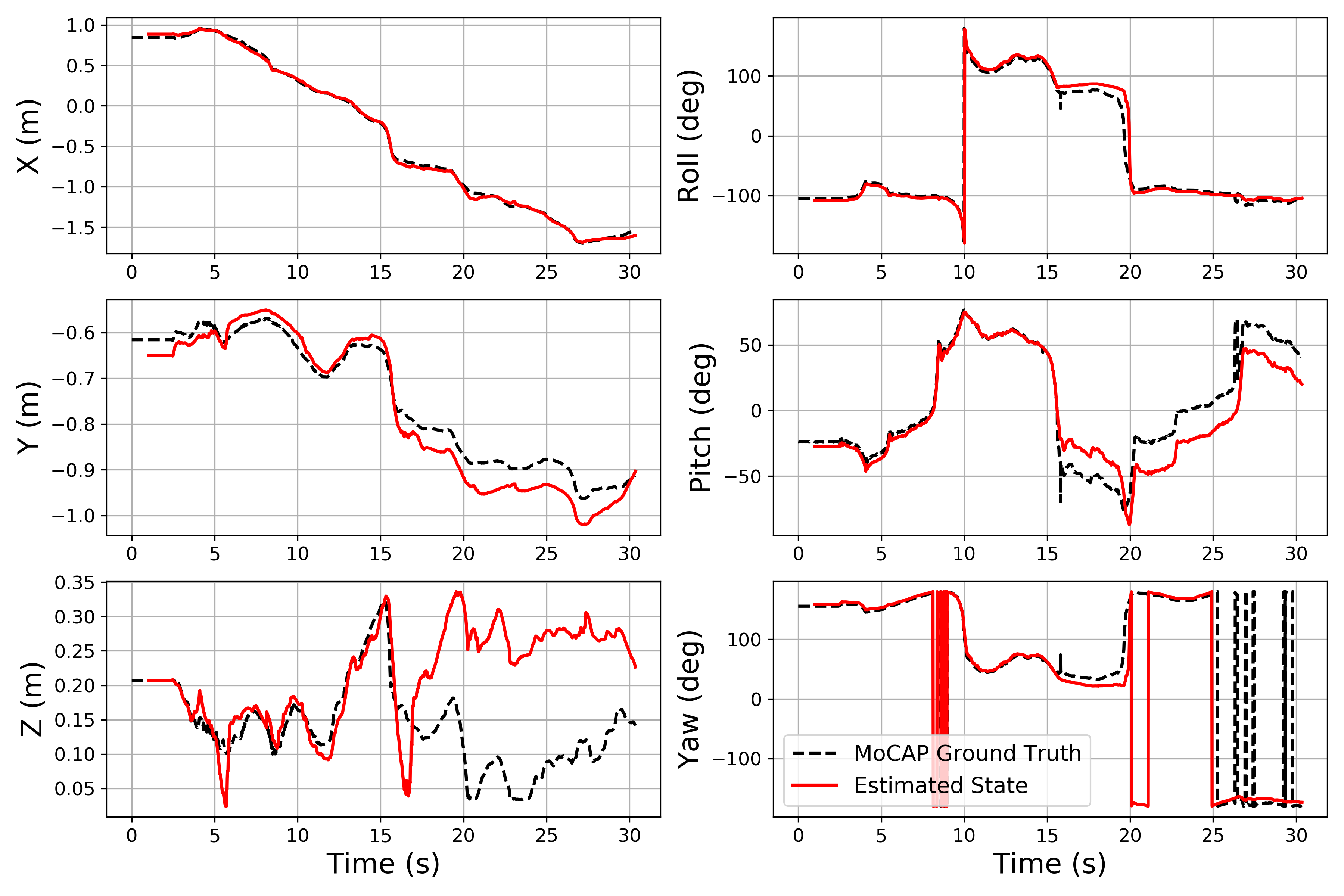}
        \caption{Estimated position and orientation vs. MoCAP ground truth.}
        \label{fig:result-realworld-exp2-pos-rpy}
        \vspace{-6mm}
    \end{figure}

    As illustrated in Fig.~\ref{fig:result-realworld-exp2-pos-rpy}, the state estimator demonstrated satisfactory performance. The x and y positions were estimated with high accuracy, whereas the z position exhibited larger deviations (0.25 m) over the 5-meter trajectory. Regarding orientation, discrepancies were observed in the pitch angle, particularly toward the latter portion of the experiment. Despite these issues, the overall performance of the estimator remains promising in the transition from simulation to real-world implementation. 
    \blue{Potential sources of error contributing to the z-axis discrepancies include the contact impact on the IMU sensor, insufficient motion excitation along the z-axis compared to the x-y plane, and drift in estimated contact points. Collectively, these factors appear to reduce the accuracy of the z-axis position estimation.}

    The root-mean-square-errors (RMSE) of the Relative Pose Error (RPE), reported in drift per meter, are summarized in Table.~\ref{tab:res-pose-error}.
    The average trajectory drift percentage (final drift/trajectory length) across all six datasets is 4.20\%.
    Given the absence of prior work on dead-reckoning or odometry for tensegrity robots, we compared our results to state-of-the-art dead-reckoning and odometry algorithms for other robotic platforms, including a final drift of 3.18\% in full-sized vehicles \cite{lin2023proprioceptive}, 1.65\% in legged robots \cite{yang2023cerberus}, and 2.38\% in the industrial robot Husky \cite{lin2023proprioceptive}. Despite the significantly increased structural and kinematic complexity of tensegrity robots, our state estimator's performance remains comparable to those of these systems.

    Runtime evaluations were performed on a personal laptop with i5-11400H CPU. Shape reconstruction required an average time of 4.1 ms, while InEKF propagation and contact correction took 11.33 µs and 17.46 µs, respectively. These results highlight the efficiency of the proposed algorithm enabling onboard computing on untethered tensegrity robots.

\section{Conclusion}
We proposed a proprioceptive contact-aided Right Invariant EKF-based state estimator for tensegrity robots. Our approach utilizes an optimization-based method to reconstruct the shape of the 3-bar triangular prism tensegrity robot, providing kinematic information to the filter for correction. The estimator can estimate the robot's pose in real-time in a real-world setting, relying solely on an IMU, motor encoder measurements, and contact information. It achieves an average positional drift percentage of 4.20\%.

This proprioceptive approach eliminates the need for off-board sensor setups while delivering pose estimates comparable in accuracy to traditional rigid robots. 
However, contact detection currently remains off-board due to hardware constraints. Furthermore, the current real-world experiments are limited in scope and do not fully showcase the state estimation capabilities across a broader range of maneuvers, such as point turning or crawling, and varied terrains.

Future work will focus on incorporating contact endcap geometry into the reconstructed kinematics to improve estimation accuracy. Additionally, we plan to investigate onboard contact detection methods, such as implementing soft endcap contact sensors based on gelslim\cite{sipos2024gelslim, donlon2018gelslim} or leveraging a graph neural network-based approach\cite{butterfield2024mi}, to enable a fully proprioceptive tensegrity robot system.


\section*{Acknowledgment}
\small{The authors thank Hybrid Dynamic Robotics Lab team member Yilin Ma's help with the experiment environment setup.}

{\footnotesize
\balance
\bibliographystyle{IEEEtran}
\bibliography{strings-abrv,ieee-abrv,ref}

\begin{thebibliography}{10}
\providecommand{\url}[1]{#1}
\csname url@samestyle\endcsname
\providecommand{\newblock}{\relax}
\providecommand{\bibinfo}[2]{#2}
\providecommand{\BIBentrySTDinterwordspacing}{\spaceskip=0pt\relax}
\providecommand{\BIBentryALTinterwordstretchfactor}{4}
\providecommand{\BIBentryALTinterwordspacing}{\spaceskip=\fontdimen2\font plus
\BIBentryALTinterwordstretchfactor\fontdimen3\font minus \fontdimen4\font\relax}
\providecommand{\BIBforeignlanguage}[2]{{%
\expandafter\ifx\csname l@#1\endcsname\relax
\typeout{** WARNING: IEEEtran.bst: No hyphenation pattern has been}%
\typeout{** loaded for the language `#1'. Using the pattern for}%
\typeout{** the default language instead.}%
\else
\language=\csname l@#1\endcsname
\fi
#2}}
\providecommand{\BIBdecl}{\relax}
\BIBdecl

\bibitem{skelton2009tensegrity}
R.~E. Skelton and M.~C. De~Oliveira, \emph{Tensegrity systems}.\hskip 1em plus 0.5em minus 0.4em\relax Springer, 2009, vol.~1.

\bibitem{shah2022tensegrity}
D.~S. Shah, J.~W. Booth, R.~L. Baines, K.~Wang, M.~Vespignani, K.~Bekris, and R.~Kramer-Bottiglio, ``Tensegrity robotics,'' \emph{Soft robotics}, vol.~9, no.~4, pp. 639--656, 2022.

\bibitem{sabelhaus2014hardware}
A.~P. Sabelhaus, J.~Bruce, K.~Caluwaerts, Y.~Chen, D.~Lu, Y.~Liu, A.~K. Agogino, V.~SunSpiral, and A.~M. Agogino, ``Hardware design and testing of superball, a modular tensegrity robot,'' in \emph{World Conference on Structural Control and Monitoring (WCSCM)}, no. ARC-E-DAA-TN15339, 2014.

\bibitem{sabelhaus2015system}
A.~P. Sabelhaus, J.~Bruce, K.~Caluwaerts, P.~Manovi, R.~F. Firoozi, S.~Dobi, A.~M. Agogino, and V.~SunSpiral, ``System design and locomotion of superball, an untethered tensegrity robot,'' in \emph{Proc. {IEEE} Int. Conf. Robot. and Automation}.\hskip 1em plus 0.5em minus 0.4em\relax IEEE, 2015, pp. 2867--2873.

\bibitem{vespignani2018design}
M.~Vespignani, J.~M. Friesen, V.~SunSpiral, and J.~Bruce, ``Design of superball v2, a compliant tensegrity robot for absorbing large impacts,'' in \emph{Proc. {IEEE}/{RSJ} Int. Conf. Intell. Robots and Syst.}\hskip 1em plus 0.5em minus 0.4em\relax IEEE, 2018, pp. 2865--2871.

\bibitem{mi2024design}
J.~Mi, W.~Tong, Y.~Ma, and X.~Huang, ``Design of a variable stiffness quasi-direct drive cable-actuated tensegrity robot,'' \emph{arXiv preprint arXiv:2409.05751}, 2024.

\bibitem{paul2006design}
C.~Paul, F.~J. Valero-Cuevas, and H.~Lipson, ``Design and control of tensegrity robots for locomotion,'' \emph{{IEEE} Trans. Robot.}, vol.~22, no.~5, pp. 944--957, 2006.

\bibitem{kim2014rapid}
K.~Kim, A.~K. Agogino, D.~Moon, L.~Taneja, A.~Toghyan, B.~Dehghani, V.~SunSpiral, and A.~M. Agogino, ``Rapid prototyping design and control of tensegrity soft robot for locomotion,'' in \emph{IEEE international conference on robotics and biomimetics}.\hskip 1em plus 0.5em minus 0.4em\relax IEEE, 2014, pp. 7--14.

\bibitem{kim2015robust}
K.~Kim, A.~K. Agogino, A.~Toghyan, D.~Moon, L.~Taneja, and A.~M. Agogino, ``Robust learning of tensegrity robot control for locomotion through form-finding,'' in \emph{Proc. {IEEE}/{RSJ} Int. Conf. Intell. Robots and Syst.}\hskip 1em plus 0.5em minus 0.4em\relax IEEE, 2015, pp. 5824--5831.

\bibitem{campos2021orb}
C.~Campos, R.~Elvira, J.~J.~G. Rodr{\'\i}guez, J.~M. Montiel, and J.~D. Tard{\'o}s, ``{ORB-SLAM3}: An accurate open-source library for visual, visual--inertial, and multimap slam,'' \emph{{IEEE} Trans. Robot.}, vol.~37, no.~6, pp. 1874--1890, 2021.

\bibitem{shan2020lio}
T.~Shan, B.~Englot, D.~Meyers, W.~Wang, C.~Ratti, and D.~Rus, ``{LIO-SAM}: Tightly-coupled lidar inertial odometry via smoothing and mapping,'' in \emph{Proc. {IEEE}/{RSJ} Int. Conf. Intell. Robots and Syst.}\hskip 1em plus 0.5em minus 0.4em\relax IEEE, 2020, pp. 5135--5142.

\bibitem{li2020dxslam}
D.~Li, X.~Shi, Q.~Long, S.~Liu, W.~Yang, F.~Wang, Q.~Wei, and F.~Qiao, ``{DXSLAM}: A robust and efficient visual slam system with deep features,'' in \emph{Proc. {IEEE}/{RSJ} Int. Conf. Intell. Robots and Syst.}\hskip 1em plus 0.5em minus 0.4em\relax IEEE, 2020, pp. 4958--4965.

\bibitem{lin2023proprioceptive}
T.-Y. Lin, T.~Li, W.~Tong, and M.~Ghaffari, ``Proprioceptive invariant robot state estimation,'' \emph{arXiv preprint arXiv:2311.04320}, 2023.

\bibitem{xu2024customizable}
X.~Xu, T.~Zhang, S.~Wang, X.~Li, Y.~Chen, Y.~Li, B.~Raj, M.~Johnson-Roberson, and X.~Huang, ``Customizable perturbation synthesis for robust slam benchmarking,'' \emph{arXiv preprint arXiv:2402.08125}, 2024.

\bibitem{chen2017soft}
L.-H. Chen, K.~Kim, E.~Tang, K.~Li, R.~House, E.~L. Zhu, K.~Fountain, A.~M. Agogino, A.~Agogino, V.~Sunspiral \emph{et~al.}, ``Soft spherical tensegrity robot design using rod-centered actuation and control,'' \emph{Journal of Mechanisms and Robotics}, vol.~9, no.~2, p. 025001, 2017.

\bibitem{hirai2013active}
S.~Hirai, Y.~Koizumi, M.~Shibata, M.~Wang, and L.~Bin, ``Active shaping of a tensegrity robot via pre-pressure,'' in \emph{IEEE/ASME International Conference on Advanced Intelligent Mechatronics}.\hskip 1em plus 0.5em minus 0.4em\relax IEEE, 2013, pp. 19--25.

\bibitem{rieffel2018adaptive}
J.~Rieffel and J.-B. Mouret, ``Adaptive and resilient soft tensegrity robots,'' \emph{Soft robotics}, vol.~5, no.~3, pp. 318--329, 2018.

\bibitem{lu20226n}
S.~Lu, W.~R. Johnson~III, K.~Wang, X.~Huang, J.~Booth, R.~Kramer-Bottiglio, and K.~Bekris, ``{6N-DoF} pose tracking for tensegrity robots,'' in \emph{Proc. Int. Symp. Robot. Res.}\hskip 1em plus 0.5em minus 0.4em\relax Springer, 2022, pp. 136--152.

\bibitem{johnson2022sensor}
W.~R. Johnson, A.~Agrawala, X.~Huang, J.~Booth, and R.~Kramer-Bottiglio, ``Sensor tendons for soft robot shape estimation,'' in \emph{IEEE Sensors}.\hskip 1em plus 0.5em minus 0.4em\relax IEEE, 2022, pp. 1--4.

\bibitem{li2021shape}
W.-Y. Li, A.~Takata, H.~Nabae, G.~Endo, and K.~Suzumori, ``Shape recognition of a tensegrity with soft sensor threads and artificial muscles using a recurrent neural network,'' \emph{IEEE Robotics and Automation Letters}, vol.~6, no.~4, pp. 6228--6234, 2021.

\bibitem{qin2018vins}
T.~Qin, P.~Li, and S.~Shen, ``Vins-mono: A robust and versatile monocular visual-inertial state estimator,'' \emph{{IEEE} Trans. Robot.}, vol.~34, no.~4, pp. 1004--1020, 2018.

\bibitem{roston1991dead}
G.~P. Roston and E.~P. Krotkov, \emph{Dead Reckoning Navigation For Walking Robots.}\hskip 1em plus 0.5em minus 0.4em\relax Department of Computer Science, Carnegie-Mellon University, 1991.

\bibitem{camurri2020pronto}
M.~Camurri, M.~Ramezani, S.~Nobili, and M.~Fallon, ``Pronto: A multi-sensor state estimator for legged robots in real-world scenarios,'' \emph{Frontiers in Robotics and AI}, vol.~7, p.~68, 2020.

\bibitem{li2013high}
M.~Li and A.~I. Mourikis, ``High-precision, consistent {EKF}-based visual-inertial odometry,'' \emph{Int. J. Robot. Res.}, vol.~32, no.~6, pp. 690--711, 2013.

\bibitem{bloesch2013state}
M.~Bloesch, M.~Hutter, M.~A. Hoepflinger, S.~Leutenegger, C.~Gehring, C.~D. Remy, and R.~Siegwart, ``State estimation for legged robots: Consistent fusion of leg kinematics and imu,'' 2013.

\bibitem{hartley2020contact}
R.~Hartley, M.~Ghaffari, R.~M. Eustice, and J.~W. Grizzle, ``Contact-aided invariant extended kalman filtering for robot state estimation,'' \emph{Int. J. Robot. Res.}, vol.~39, no.~4, pp. 402--430, 2020.

\bibitem{yu2023fully}
X.~Yu, S.~Teng, T.~Chakhachiro, W.~Tong, T.~Li, T.-Y. Lin, S.~Koehler, M.~Ahumada, J.~M. Walls, and M.~Ghaffari, ``Fully proprioceptive slip-velocity-aware state estimation for mobile robots via invariant {Kalman} filtering and disturbance observer,'' in \emph{Proc. {IEEE}/{RSJ} Int. Conf. Intell. Robots and Syst.}\hskip 1em plus 0.5em minus 0.4em\relax IEEE, 2023, pp. 8096--8103.

\bibitem{xiong2019imu}
L.~Xiong, X.~Xia, Y.~Lu, W.~Liu, L.~Gao, S.~Song, Y.~Han, and Z.~Yu, ``{IMU}-based automated vehicle slip angle and attitude estimation aided by vehicle dynamics,'' \emph{Sensors}, vol.~19, no.~8, p. 1930, 2019.

\bibitem{tietz2013tetraspine}
B.~R. Tietz, R.~W. Carnahan, R.~J. Bachmann, R.~D. Quinn, and V.~SunSpiral, ``Tetraspine: Robust terrain handling on a tensegrity robot using central pattern generators,'' in \emph{IEEE/ASME International Conference on Advanced Intelligent Mechatronics}.\hskip 1em plus 0.5em minus 0.4em\relax IEEE, 2013, pp. 261--267.

\bibitem{huang2022live}
X.~Huang, W.~R. Johnson, J.~Booth, and R.~Kramer-Bottiglio, ``Live demonstration: Tensegrity state estimation,'' in \emph{IEEE Sensors}.\hskip 1em plus 0.5em minus 0.4em\relax IEEE, 2022, pp. 1--1.

\bibitem{booth2021surface}
J.~W. Booth, O.~Cyr-Choiniere, J.~C. Case, D.~Shah, M.~C. Yuen, and R.~Kramer-Bottiglio, ``Surface actuation and sensing of a tensegrity structure using robotic skins,'' \emph{Soft Robotics}, vol.~8, no.~5, pp. 531--541, 2021.

\bibitem{moldagalieva2019computer}
A.~Moldagalieva, D.~Fadeyev, A.~Kuzdeuov, V.~Khan, B.~Alimzhanov, and H.~A. Varol, ``Computer vision-based pose estimation of tensegrity robots using fiducial markers,'' in \emph{IEEE/SICE International Symposium on System Integration (SII)}.\hskip 1em plus 0.5em minus 0.4em\relax IEEE, 2019, pp. 478--483.

\bibitem{caluwaerts2016state}
K.~Caluwaerts, J.~Bruce, J.~M. Friesen, and V.~SunSpiral, ``State estimation for tensegrity robots,'' in \emph{Proc. {IEEE} Int. Conf. Robot. and Automation}.\hskip 1em plus 0.5em minus 0.4em\relax IEEE, 2016, pp. 1860--1865.

\bibitem{barrau2015ekf}
A.~Barrau and S.~Bonnabel, ``An {EKF-SLAM} algorithm with consistency properties,'' \emph{arXiv preprint arXiv:1510.06263}, 2015.

\bibitem{barrau2015non}
A.~Barrau, ``Non-linear state error based extended {Kalman} filters with applications to navigation,'' Ph.D. dissertation, Mines Paristech, 2015.

\bibitem{luo2020geometry}
Y.~Luo, M.~Wang, and C.~Guo, ``The geometry and kinematics of the matrix {Lie} group $ se\_k (3) $,'' \emph{arXiv preprint arXiv:2012.00950}, 2020.

\bibitem{grupp2017evo}
M.~Grupp, ``evo: Python package for the evaluation of odometry and slam.'' \url{https://github.com/MichaelGrupp/evo}, 2017.

\bibitem{yang2023cerberus}
S.~Yang, Z.~Zhang, Z.~Fu, and Z.~Manchester, ``Cerberus: Low-drift visual-inertial-leg odometry for agile locomotion,'' in \emph{Proc. {IEEE} Int. Conf. Robot. and Automation}.\hskip 1em plus 0.5em minus 0.4em\relax IEEE, 2023, pp. 4193--4199.

\bibitem{sipos2024gelslim}
A.~Sipos, W.~v.~d. Bogert, and N.~Fazeli, ``{GelSlim 4.0}: Focusing on touch and reproducibility,'' \emph{arXiv preprint arXiv:2409.19770}, 2024.

\bibitem{donlon2018gelslim}
E.~Donlon, S.~Dong, M.~Liu, J.~Li, E.~Adelson, and A.~Rodriguez, ``Gelslim: A high-resolution, compact, robust, and calibrated tactile-sensing finger,'' in \emph{Proc. {IEEE}/{RSJ} Int. Conf. Intell. Robots and Syst.}\hskip 1em plus 0.5em minus 0.4em\relax IEEE, 2018, pp. 1927--1934.

\bibitem{butterfield2024mi}
D.~Butterfield, S.~S. Garimella, N.-J. Cheng, and L.~Gan, ``{MI-HGNN}: Morphology-informed heterogeneous graph neural network for legged robot contact perception,'' \emph{arXiv preprint arXiv:2409.11146}, 2024.

\end{thebibliography}
}

\end{document}